\documentclass[journal]{IEEEtran}
\usepackage{graphicx}
\usepackage{cite}
\usepackage{picinpar}
\usepackage{url}
\usepackage[latin1]{inputenc}
\usepackage{colortbl}
\usepackage{soul}
\soulregister\cite7
\soulregister\ref7
\usepackage{multirow}
\usepackage{pifont}
\usepackage{color}
\usepackage{alltt}
\usepackage[hidelinks]{hyperref}
\usepackage{enumerate}
\usepackage{siunitx}
\usepackage{breakurl}
\usepackage{epstopdf}
\usepackage{pbox}
\usepackage{amsmath,amssymb,amsfonts}
\usepackage{textcomp}
\usepackage{wrapfig}
\usepackage{empheq}
\usepackage{array}
\usepackage{makecell}
\usepackage{xcolor}
\usepackage{booktabs}
\usepackage{threeparttable}
\usepackage{adjustbox}
\usepackage{subfigure}
\usepackage{verbatim}
\usepackage{gensymb}
\usepackage[misc,geometry]{ifsym}
\graphicspath{{figures/}}
\definecolor{ccr}{RGB}{10,50,200}
\usepackage{hyperref}
\hypersetup{hypertex=true,
	colorlinks=true,
	linkcolor=ccr,
	anchorcolor=ccr,
	citecolor=ccr}
\usepackage{multicol}
\usepackage[algoruled,vlined,linesnumbered]{algorithm2e}
\usepackage{caption}
\begin{document}
	\title{Hierarchical Diffusion Policy: manipulation trajectory generation via contact guidance}
	
	\author{
		\vskip 1em
		
		Dexin Wang, 
		Chunsheng Liu, \emph{Member, IEEE},
		Faliang Chang, 
		Yichen Xu,
		
		\thanks{
			All authors are with the School of Control Science and Engineering, Shandong University, Ji'nan, Shandong 250061, China (e-mail: flchang@sdu.edu.cn, liuchunsheng@sdu.edu.cn). (Corresponding author: Faliang Chang, Chunsheng Liu)
		}
	}
	
	\maketitle
	
	\begin{abstract}
		Decision-making in robotics using denoising diffusion processes has increasingly become a hot research topic, but end-to-end policies perform poorly in tasks with rich contact and have limited controllability.
		This paper proposes Hierarchical Diffusion Policy (HDP), a new imitation learning method of using objective contacts to guide the generation of robot trajectories.
		The policy is divided into two layers: the high-level policy predicts the contact for the robot's next object manipulation based on 3D information, while the low-level policy predicts the action sequence toward the high-level contact based on the latent variables of observation and contact.
		We represent both-level policies as conditional denoising diffusion processes, and combine behavioral cloning and Q-learning to optimize the low-level policy for accurately guiding actions towards contact.
		We benchmark Hierarchical Diffusion Policy across 6 different tasks and find that it significantly outperforms the existing state-of-the-art imitation learning method Diffusion Policy with an average improvement of 20.8\%.
		We find that contact guidance yields significant improvements, including superior performance, greater interpretability, and stronger controllability, especially on contact-rich tasks.
		To further unlock the potential of HDP, this paper proposes a set of key technical contributions including snapshot gradient optimization, 3D conditioning, and prompt guidance, which improve the policy's optimization efficiency, spatial awareness, and controllability respectively.
		Finally, real-world experiments verify that HDP can handle both rigid and deformable objects.
		Code, data and training details can be found at \href{https://github.com/dexin-wang/Hierarchical_Diffusion_Policy}{https://github.com/dexin-wang/Hierarchical\_Diffusion\_Policy}.
	\end{abstract}
	

	\section{Introduction}
	
	A learning-based robotic manipulation policy can be represented as a map of observations to optimal actions to accomplish a task.
	However, compared to other learning problems, predicting robot actions is inherently challenging due to factors such as 3D scene perception, rich contact, and human-robot interaction.
	
	Previous researchers attempts to adopt different learning strategies to address this challenge.
	Reinforcement learning has been widely used to learn robotics tasks since it surpassed humans in playing Atari games \cite{mnih2015human}.
	The exploration mechanism enables the agent to learn complex operational skills, but also brings low sample efficiency and weak generalization \cite{ladosz2022exploration}.
	In contrast, imitation learning learns solely from demonstrations, which significantly improves sample efficiency and generalization. 
	However, it struggles to achieve performance beyond that of the expert.
	Meanwhile, most imitation learning methods struggle to handle multimodal action distributions because they are modeled as unimodal policies \cite{zheng2024imitation}.
	Recently, diffusion models have attracted attention due to their ability to model complex data distributions \cite{song2020denoising}\cite{nichol2021improved}\cite{rombach2022high}.
	Some researchers have tried to introduce diffusion models to generate robot actions \cite{wang2022diffusion}\cite{janner2022planning}.
	Chi \textit{et al.} proposed Diffusion policy to jointly reason about multi-step action sequences, which achieved significant improvements \cite{chi2023diffusion}. 
	Diffusion policy can express multi-modal action distribution and cope with high-dimensional output space while maintaining high sample efficiency.
	However, Diffusion policy performs worse in the contact-rich tasks than in tasks with single contact.
	We argue that the reason for this phenomenon is that the Diffusion policy does not explicitly model the interaction between the robot and object.
	In this paper, we introduce contact position to model the interaction between the robot and object.
	Specifically, we design a new network to predict the objective contact position of the robot and utilize reinforcement learning to guide the policy to generate robot trajectories toward the objective contact.
	
	\begin{figure}[tp]
		\centering
		{\includegraphics[scale=0.28]{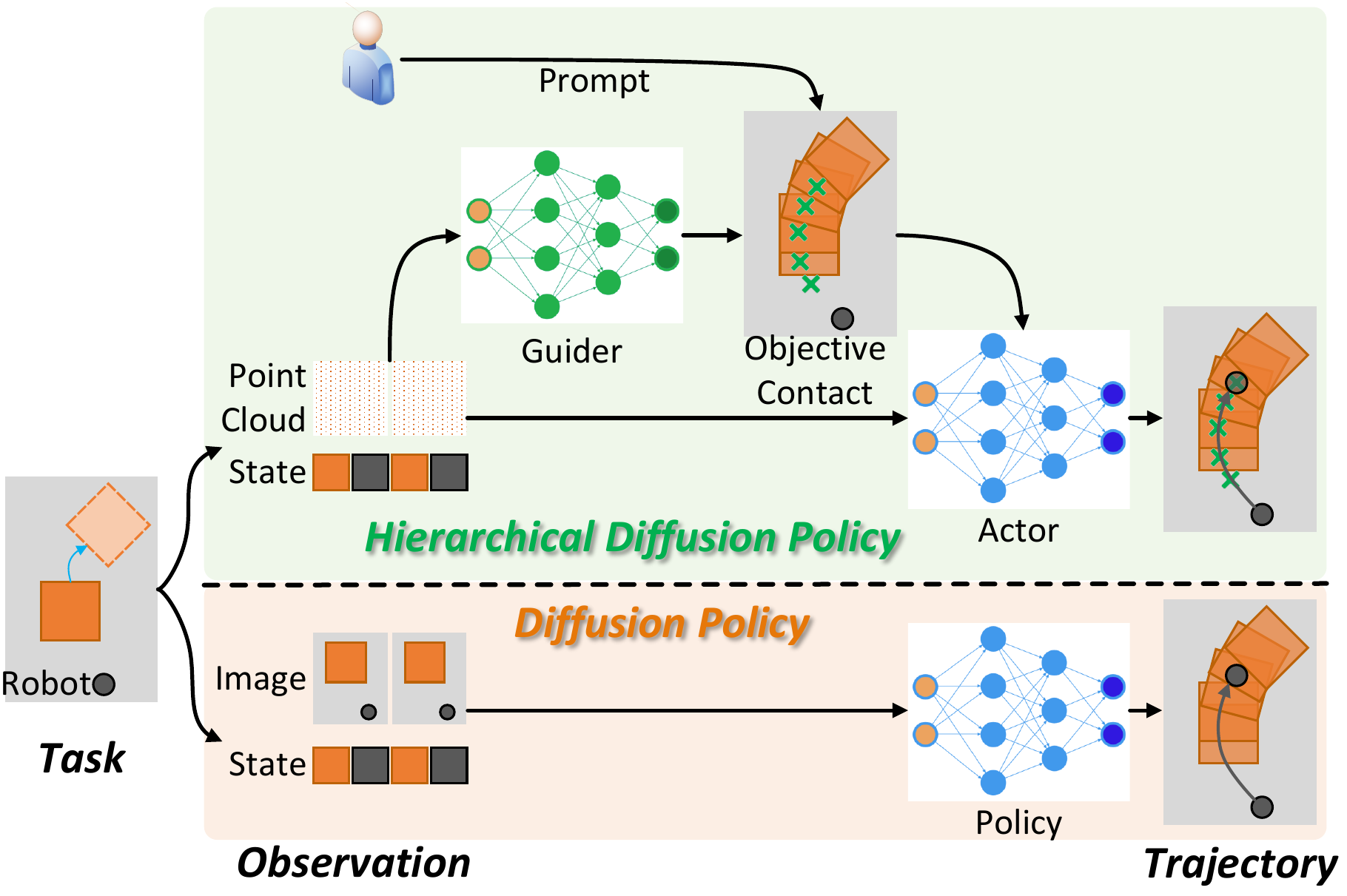}}
		\caption{
			\textbf{Inference Process of Hierarchical Diffusion Policy}.
			Compared to the Diffusion Policy that generates operation trajectories end-to-end, HDP introduces objective contacts, predicted by the Guider network or provided by humans, to guide the trajectory generation.
		}
		\label{fig_introduction}
	\end{figure}
	
	In this paper, we propose \textbf{Hierarchical Diffusion Policy} (\textbf{HDP}), a novel method that uses contact to guide the generation of robot manipulation trajectories (Fig.~\ref{fig_introduction}).
	HDP consists of three networks, namely Guider, Actor, and Critic, which predict objective contact, robot action sequence, and Q-value respectively.
	The Guider predicts the contact through a conditional denoising diffusion process, which represents the operating position (already in contact) or the contact position to reach the object (not yet in contact).
	The Actor is also modeled as a conditional denoising diffusion model, which predicts action sequences based on historical observations and objective contacts.
	We jointly optimize the Actor using behavior cloning and Q-learning algorithms to enable the Actor to establish the guidance of actions by contacts, while learning the conditional probability distribution of actions given observations.
	By introducing contact guidance for actions, HDP divides the manipulation planning into high-level contact planning and low-level trajectory generation, which brings three advantages:
	
	\begin{itemize}
		\item \textbf{Superior performance}.
		Short-horizon operation tasks are easier to learn than long-horizon tasks, and the contact condition makes the Actor focus more on short-horizon operation planning.
		We build the Guider based on the diffusion process to express multimodal contact distributions, thereby achieving accurate high-level planning.
		
		\item \textbf{Greater interpretability}.
		Hierarchical planning makes the internal logic of the trajectory generated at each step more transparent.
		Thanks to the precise guidance of robot actions towards objective contact, the robot's operational intent becomes apparent well in advance.
		
		\item \textbf{Stronger controllability}.
		We can intervene in the operation process at any time and guide the robot to complete the task in the desired way by designating the objective contact positions, which enhances the human-machine interaction.
	\end{itemize}
	
	To further unlock the potential of HDP, this paper proposes several key technical contributions.
	
	\begin{itemize}
		\item \textbf{Snapshot gradient optimization}.
		Wang \textit{et al.} optimize the diffusion-based policy by calculating the gradient of the policy at all timesteps along the diffusion process in the Q-learning algorithm, which resulted in inefficient training \cite{wang2022diffusion}.
		We propose the snapshot gradient optimization strategy that calculates the gradient of only one timestep in the diffusion process per iteration, significantly improving training speed.
		
		\item \textbf{3D conditioning}.
		To improve the perception of the policy, we introduce 3D point clouds to represent the state.
		The policy extracts the 3D representation once regardless of the denoising iteration.
		
		\item \textbf{Prompt guidance}.
		Manually designating objective contacts without using the Guider can produce customized operation trajectories.
		This demonstrates the potential application of HDP in enhancing human-machine interaction.
	\end{itemize}
	
	We systematically evaluate HDP across 6 tasks.
	The evaluation includes both simulated and real-world environments, 2DOF to 6DOF actions, prehensile and non-prehensile operations, with rigid and deformable objects, using demonstration data collected by single and multiple users.
	Empirically, we find significant performance boost with an average improvement of 20.8\%, providing strong evidence of the effectiveness of HDP. 
	We also provide detailed analysis to carefully examine the characteristics of the proposed algorithm and the impacts of the key design decisions.
	The code, data, and training details will be publicly available.

	\section{Related work}
	Long-horizon tasks have been a hot topic in the field of robotics in recent years due to their complexity and sparse rewards \cite{guo2023recent}\cite{orthey2023sampling}.
	An intuitive approach is to hierarchically decompose long-horizon tasks to reduce task complexity \cite{triantafyllidis2023hybrid}.
	From a learning perspective, long-horizon tasks are difficult to learn using reinforcement learning algorithms due to sparse rewards, resulting in very low training efficiency.
	Additionally, frequent interactions can increase the safety risks for the robot \cite{singh2022reinforcement}.
	Current research trends are gradually shifting towards learning manipulation policies from pre-collected demonstrations \cite{zheng2024imitation}\cite{prudencio2023survey}.
	Below we briefly review related work on hierarchical manipulation and learning from demonstrations.

	\subsection{Hierarchical Manipulation}
	Non-learning methods usually decompose the complete state space and perform continuous planning separately.
	A representative approach first searches for the object's movement trajectory in space, then searches for the robot's operational trajectory, and finally optimizes the results \cite{aceituno2022hierarchical}\cite{cheng2023enhancing}.
	This type of method is stable but limited to objects and scenes with known models, resulting in insufficient generalization.
	
	Most learning-based methods are categorized by the scope of high-level actions into task-specific, finite set, LLM-based planning, fixed-step planning, and object motion-related planning.
	Task-specific methods limit high-level actions to specific tasks related to the task, and are highly efficient in learning but cannot be used for other tasks \cite{simonivc2024hierarchical}\cite{9745587}\cite{luo2024multi}.
	Finite set-based methods collect reusable skills from demonstrations as high-level actions, which are also highly efficient in learning but can only be used for a few tasks composed of skills \cite{zhu2022bottom}\cite{yang2021hierarchical}\cite{rana2023residual}.
	With the rapid development of LLMs, some methods use the commonsense knowledge of LLMs for task planning. 
	While they can handle most everyday tasks, they struggle with tasks requiring precise manipulation \cite{ajay2024compositional}\cite{jia2023chain}\cite{brohan2023can}.
	Hierarchical reinforcement learning methods, which predict high-level actions with fixed steps, are widely used in various types of robots, but they do not focus on the objects to be manipulated \cite{hafner2022deep}\cite{zhang2022adjacency}\cite{huang2022planning}.
	Ma \textit{et al.} decompose the prehensile task into predicting the grasping posture and predicting the robot trajectory, which effectively improved the success rate \cite{ma2024hierarchical}.
	In contrast, our previous work \cite{wang2024multi} use contact as a subgoal of the low-level policy, which can be easily extended to different end-effectors (such as dexterous hands and single-finger grippers) rather than being limited to parallel-jaw grippers. 
	Additionally, it can be applied to non-prehensile tasks (such as flipping and tilting).
	In this paper, we improve the contact planning in our previous work \cite{wang2024multi} from being computed based on an object model to being generated by a denoised diffusion process to improve generalization.

	\subsection{Learning from Demonstrations}
	The simplest form of learning from demonstration is behavioral cloning, which is copying the behavioral policy of collecting demonstrations.
	Current behavior cloning approaches can be categorized into two groups, depending on the policy's structure.
	Explicit policies directly map observations to actions \cite{zhang2018deep}\cite{florence2019self}\cite{toyer2020magical}.
	To model multimodal action distributions, some methods model the mapping as a classification problem instead of a regression problem \cite{zeng2021transporter}\cite{avigal2022speedfolding}.
	Further approaches combine categorical and Gaussian distributions to represent continuous multimodal distributions \cite{mandlekar2022matters}.
	However, these models often face challenges such as difficult hyperparameter tuning, model collapse, and limited expression \cite{florence2022implicit}.
	Implicit policies use energy-based models to model the action distribution, which can effectively express multimodal distribution \cite{du2021improved}\cite{grathwohl2020learning}.
	However, existing implicit policies \cite{florence2022implicit} are unstable to train due to the necessity of drawing negative samples.
	
	One of the main issues with behavior cloning is its inability to learn policies that surpass the behaviors demonstrated in the collected data.
	A solution to this problem is offline reinforcement learning, which optimizes policies through value functions, allowing for better handling of suboptimal data and adaptation to unseen states \cite{zhan2022offline}\cite{cheng2022adversarially}\cite{luo2023action}.
	Unlike online reinforcement learning, offline reinforcement learning cannot correct erroneous behaviors by interacting with the environment, so it is necessary to find a balance between increased generalization and avoiding unwanted behaviors outside of distribution, namely distributional shift \cite{prudencio2023survey}.
	Many researchers have proposed different solutions to distribution shifts, including policy constraints \cite{fujimoto2019off}\cite{nair2020awac}, importance sampling \cite{precup2000eligibility}\cite{zhanggendice}, regularization \cite{haarnoja2018soft}\cite{kumar2020conservative}, uncertainty estimation \cite{agarwal2020optimistic}, etc.
	We combine offline reinforcement learning with behavior cloning, and the optimization objectives include minimizing the action difference and maximizing the value function, corresponding to the above-mentioned policy constraints.

	\subsection{Diffusion Models}
	The diffusion model, also known as the denoised diffusion probability model, generates data that conforms to a specific distribution through an iterative denoising process.
	They have demonstrated amazing generative capabilities on both unconditional and conditional image, video, and 3D object generation \cite{rombach2022high}\cite{ho2022cascaded}\cite{dhariwal2021diffusion}\cite{wu2023tune}\cite{liu2024sora}\cite{vahdat2022lion}\cite{luo2021diffusion}.
	In the decision-making domain, diffusion models have also recently been used to model multimodal action distributions \cite{song2020denoising}\cite{nichol2021improved}\cite{rombach2022high}\cite{black2023training}\cite{ajayconditional}.
	Among them, Diffusion Policy constructs a diffusion-based robot trajectory generation policy with behavior cloning \cite{chi2023diffusion}.
	Some researchers have enhanced Diffusion Policy from various aspects, such as state reconstruction \cite{li2023crossway} and hierarchical policy \cite{ma2024hierarchical}.
	
	Concurrent with our work, Ma et al. proposed to first predict the robot's end pose and then predict the action sequence ending at the target pose through a diffusion-based policy with behavior cloning \cite{ma2024hierarchical}.
	However, due to the limitations of behavior cloning, they still face the same issues as Diffusion Policy, namely the difficulty in learning policies beyond the behavior policy and the accumulation of errors.
	In addition, pose is not sufficient to express the states of different end-effectors, such as dexterous hands with more contact points.
	In contrast, our method introduces an additional value function that enables the policy to focus on longer time steps rather than just the current state, allowing for better handling of suboptimal data and adaptation to unseen states.
	Additionally, we use contact states as high-level actions, which makes it easier to apply to end-effectors with different numbers of fingers.

	\section{Hierarchical Diffusion Policy Formulation}
	
	We formulate HDP as Conditional Denoising Diffusion Probabilistic Models (CDDPMs) \cite{ho2022classifier}.
	HDP is able to express multimodal contact distributions and predict action sequences that are also multimodal based on contact conditions.
	Crucially, HDP can ensure the controllability of trajectory generation.
	The following sections describe CDDPMs in more detail and explain how construct HDP.

	\subsection{Denoising Diffusion Probabilistic Models}
	\label{DDPMs}
	
	Denoising Diffusion Probabilistic Models (DDPMs) are a type of generative model that generates high-quality data samples by gradually denoising \cite{nichol2021improved}.
	It includes a forward process and a reverse process.
	
	The forward process begins with the action data $x_0$ and defines a Markov chain that progressively adds noise over $K$ steps:
	\begin{align}
		q(x^k|x^{k-1}) = \mathcal{N}(x^k; \sqrt{\alpha^k} x^{k-1}, (1-\alpha^k)\mathbf{I})
		\label{eq_31}
	\end{align}
	where $\alpha^k$ is the noise scheduling parameter at timestep $k$.
	The iterative noise addition process can be combined into the following formula:
	\begin{align}
		x^k = \sqrt{\bar{\alpha}^k} x^0 + \sqrt{1-\bar{\alpha}^k} z^k
		\label{eq_32}
	\end{align}
	where $\bar{\alpha}^k=\begin{matrix} \prod_{i=1}^K \alpha_i \end{matrix}$, $z^k$ is the random normally distributed noise at step $k$.
	
	The reverse process starts with $x^K$ sampled from Gaussian noise, the DDPM undergoes $K$ iterations of denoising, generating a sequence of intermediate actions with progressively reduced noise levels, until achieving the desired noise-free output $x^0$.
	The process follows the equation:
	\begin{align}
		p_{\theta}(x^{k-1}|x^k, y) & = 
		\notag
		\\ \mathcal{N}( x^{k-1}; & \frac{1}{\sqrt{\alpha^k}}(x^k-\frac{1-\alpha^k}{\sqrt{1-\bar{\alpha}^k}} \epsilon_{\theta}(x^k, k, y) ), (1-\alpha^k) \mathbf{I} )
		\label{eq_33}
	\end{align}
	where $y$ represents the conditional information considered during data generation. 
	$\epsilon_\theta$ is the noise prediction network with parameters $\theta$, which will be optimized through the learning process, and the learning target is $z^k$.
	
	By transforming Eq.~\ref{eq_32}, $x^0$ can be calculated directly from $x^k$ without the need for iterative reverse diffusion, as follows:
	\begin{align}
		x^0 = \frac{1}{\sqrt{\bar{\alpha}^k}}(x^k - \sqrt{1-\bar{\alpha}^k} z^k)
		\label{eq_34}
	\end{align}

	\begin{figure*}[tp]
		\centering
		{\includegraphics[scale=0.45]{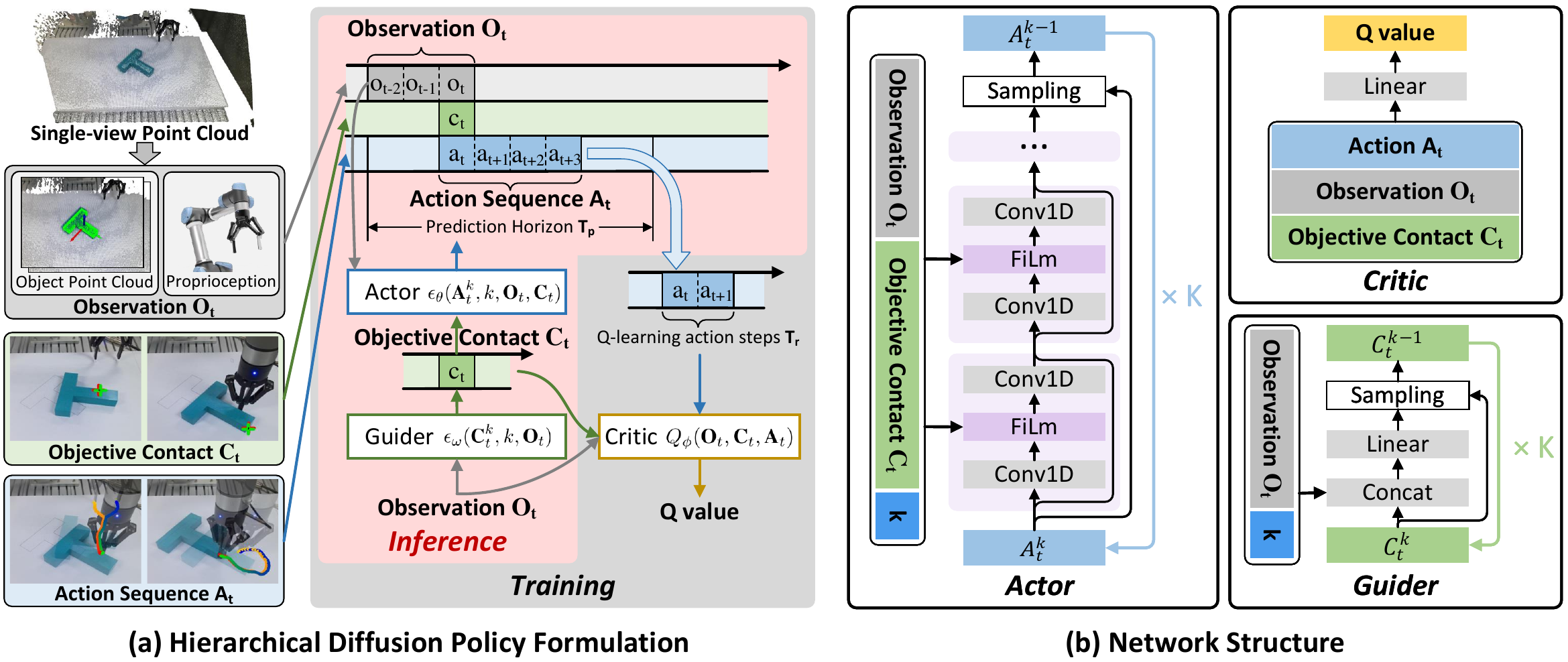}}
		\caption{
			\textbf{Hierarchical Diffusion Policy Overview.}
			\textbf{(a)} At time step $t$ during inference, the Guider takes the latest $T_o$ steps of observation data $\mathbf{O_t}$ as input and predicts objective contact $\mathbf{C_t}$, the Actor  takes observation data $\mathbf{O_t}$ and objective contact $\mathbf{C_t}$ as input and predicts $T_p$ steps of actions, of which $T_a$ steps of actions are executed on the robot without re-planning.
			During training, in addition to minimizing the prediction error of the Actor, Guider, and Critic compared to the ground truth, the Actor's weights are also optimized by maximizing the Q-values.
			\textbf{(b)} The Actor and Guider are modeled as conditional denoising diffusion models, with their networks built on one-dimensional convolutions and linear layers, respectively. The Actor's architecture is similar to that of the Diffusion Policy.
			Critic is built based on linear layers.
		}
		\label{fig_overview}
	\end{figure*}

	\subsection{Hierarchical Policy Learning}
	A single model may be able to simultaneously output plans and actions at different levels, but it is difficult to guide actions using plans.
	We draw inspiration from hierarchical reinforcement learning by using different models to predict decisions at various levels and guiding actions through Q-learning.
	
	\textbf{Guider.}
	We observe that the robot can move freely before making contact with the object, but the contact positions that successfully complete the task follow a certain distribution.
	Therefore, we propose using the expected contact position between the robot and the object at time step $t$ as the high-level planning goal, denoted as $\textbf{C}_t$.
	Like the operation trajectory, the distribution of contact positions is also multimodal, meaning the robot can complete the task by contacting and manipulating different positions on the object.
	Due to the advantages of diffusion models in expressing multimodal distributions, we represent the prediction of contact positions as a conditional denoising diffusion probabilistic process, conditioned on the observation $\textbf{O}_t$.
	We construct the Guider based on a multilayer perceptron, as shown in Fig.~\ref{fig_overview}.
	Due to the 3D attributes of contact positions, we include object point clouds in the observations to enhance the network's spatial perception, along with proprioceptive information from the robot.
	To capture the conditional distribution  $p(\textbf{C}_t| \textbf{O}_t)$, we modify Eq.~\ref{eq_33} to:
	\begin{align}
		p_\omega(\textbf{C}^{k-1}_t | & \textbf{C}^{k}_t, \textbf{O}_t)  = 
		\notag
		\\ \mathcal{N}( \textbf{C}^{k-1}_t; & \frac{1}{\sqrt{\alpha^k}}(\textbf{C}^{k}_t-\frac{1-\alpha^k}{\sqrt{1-\bar{\alpha}^k}} \epsilon_\omega(\textbf{C}^{k}_t, k, \textbf{O}_t) ), (1-\alpha^k) \mathbf{I} )
		\label{eq_41}
	\end{align}
	where $\epsilon_\omega$ is the Guider network with parameters $\omega$.
	The training loss is formulated as:
	\begin{align}
		\mathcal{L}_g = MSE(z^k, \epsilon_\omega(\textbf{C}^k_t, k, \textbf{O}_t))
		\label{eq_42}
	\end{align}
	The details of the 3D encoder design are in Sec.~\ref{sec_3d_encoder}, and the details of constructing the Guider's generation targets are in Sec.~\ref{sec_oc}.
	
	\textbf{Actor.}
	We use Diffusion policy as the baseline for our new policy \cite{chi2023diffusion}, which integrates the action-sequence prediction produced by a diffusion model with receding horizon control \cite{mayne1988receding} to achieve robust action execution.
	While Diffusion policy is used to predict the robot's trajectory, we want the trajectory $\textbf{A}_t$ generated by the policy to be towards the objective contact.
	There are two major changes in the network structure: (1) using object point clouds and robot proprioception as the observation $\textbf{O}_t$; (2) using the objective contact $\textbf{C}_t$ as a generation condition other than observation.
	
	To capture the conditional distribution  $p(\textbf{A}_t| \textbf{O}_t, \textbf{C}_t)$, we modify Eq.~\ref{eq_33} to:
	\begin{align}
		p_\theta(\textbf{A}^{k-1}_t | & \textbf{A}^{k}_t, \textbf{O}_t, \textbf{C}_t)  = \mathcal{N}( \textbf{A}^{k-1}_t; 
		\notag
		\\ \frac{1}{\sqrt{\alpha^k}} & (\textbf{A}^{k}_t-\frac{1-\alpha^k}{\sqrt{1-\bar{\alpha}^k}} \epsilon_{\theta}(\textbf{A}^{k}_t, k, \textbf{O}_t, \textbf{C}_t) ), (1-\alpha^k) \mathbf{I} )
		\label{eq_43}
	\end{align}
	where $\epsilon_\theta$ is the Actor network with parameters $\theta$.
	So far, the preliminary loss function can be formulated as:
	\begin{align}
		\mathcal{L}_a = MSE(z^k, \epsilon_\theta(\textbf{A}^k_t, k, \textbf{O}_t, \textbf{C}_t))
		\label{eq_44}
	\end{align}

	\textbf{Critic.}
	Although the conditions include contact positions, the policy struggles to generate optimal trajectories that move towards or remain at the objective contacts to manipulate the object, because the loss function in Eq.~\ref{eq_44} does not establish an explicit optimization relationship between the trajectory and the objective contact positions.
	Inspired by literature \cite{wang2022diffusion}, we introduce Q-learning to establish this optimization relationship.
	They optimize the policy by using Q-learning to maximize the Q-value of the generated trajectory with respect to the entire task.
	However, complex operating tasks lead to low sample efficiency.
	
	In this work, we decompose the entire manipulation task into consecutive short-term subtasks using the objective contacts, and then maximize the Q-value of generated trajectories with respect to the subtasks.
	Specifically, the optimization goal of the Actor is to minimize the error of the predicted noise predicted and maximize the Q-value of the generated trajectory relative to the objective contact.
	The modified loss function of the Actor is as follows:
	\begin{align}
		\mathcal{L}_a = MSE(z^k, \epsilon_\theta(\textbf{A}^k_t, k, \textbf{O}_t, \textbf{C}_t)) - \boldsymbol{\eta} Q_\phi(\textbf{S}_t, \textbf{C}_t, \textbf{A}_{t, \theta}^0)
		\label{eq_45}
	\end{align}
	where $\boldsymbol{\eta}$ is weight coefficients, $\textbf{A}_{t, \theta}^0$ is the trajectory obtained after denoising $\textbf{A}^k_t$ using the Actor, $Q_\phi$ is the Critic network with parameters $\phi$. 
	We retain the original MSE loss of the Actor to enhance robustness and optimization efficiency.
	We discuss the impact of the Critic on HDP's performance in detail in Sec.~\ref{sec_q_learning}.
	
	Based on the theory of DDPMs in Sec.~\ref{DDPMs}, there are two ways for the Actor to generate $\textbf{A}_{t, \theta}^0$: iterative denoising and oneshot denoising.
	Following Eq.~\ref{eq_33}, iterative denoising can be expressed as:
	\begin{align}
		\textbf{A}_{t, \theta}^{k-1} = \frac{1}{\sqrt{\alpha^k}}(\textbf{A}^{k}_t-\frac{1-\alpha^k}{\sqrt{1-\bar{\alpha}^k}} \epsilon_{\theta}(\textbf{A}^{k}_t, k, \textbf{O}_t, \textbf{C}_t) ) 
		\notag
		\\ + \mathcal{N}(0, (1-\alpha^k) \mathbf{I})
		\label{eq_46}
	\end{align}
	Following Eq.~\ref{eq_34}, oneshot denoising can be expressed as:
	\begin{align}
		\textbf{A}_{t, \theta}^0 = \frac{1}{\sqrt{\bar{\alpha}^k}}(\textbf{A}_t^k - \sqrt{1-\bar{\alpha}^k} \epsilon_\theta(\textbf{A}^k_t, k, \textbf{O}_t, \textbf{C}_t))
		\label{eq_47}
	\end{align}
	Iterative denoising means that the optimizer will jointly compute the gradients of the weights at all time steps, while oneshot denoising only computes the gradients at time step $k$.
	Detailed discussion on these two methods is in Sec.~\ref{sec_q_learning}.
	
	Critic is built based on a multilayer perceptron, shown in Fig.~\ref{fig_overview}.
	According to the original Q-learning algorithm, the Critic's loss function can be written as follows:
	\begin{align}
		\mathcal{L}_c = MSE(r+\gamma Q_{\phi'}(\textbf{O}_{t+T_r}, \textbf{C}'_{t+T_r} & , \textbf{A}_{t+T_r, \theta}), 
		\notag
		\\ & Q_{\phi}(\textbf{O}_t, \textbf{C}_t, \textbf{A}_t))
		\label{eq_48}
	\end{align}
	where $r$ is reward. If the robot reaches the objective contact $\textbf{C}_t$ within the first $T_r$ steps of executing the predicted trajectory $A_t$, $r$ is set to a constant $R$; otherwise, it is set to 0.
	$\gamma$ is the discount factor, $Q_{\phi'}$ is target Critic network, $A_{t+T_r, \theta}$ is the trajectory predicted by the Actor at time step $t+T_r$.
	$\textbf{C}'_{t+T_r}$ is the objective contact, and its position relative to the object is the same as $\textbf{C}_t$.
	However, the demonstration used for training in this paper does not contain the data tuple $(\textbf{O}_{t+T_r}, \textbf{C}'_{t+T_r}, \textbf{A}_{t+T_r, \theta})$, making it difficult for training to converge.
	Instead we use the data tuple from the demonstration and modify the Critic's loss function to the following:
	\begin{align}
		\mathcal{L}_c = MSE(r+\gamma Q_{\phi'}(\textbf{O}_{t+T_r}, \textbf{C}'_{t+T_r} & , \textbf{A}_{t+T_r}), 
		\notag
		\\ & Q_{\phi}(\textbf{O}_t, \textbf{C}_t, \textbf{A}_t))
		\label{eq_49}
	\end{align}
	
	Based on the constructed formulation, HDP is trained as follows:
	(1) sampling data tuples $(\textbf{O}_t, \textbf{C}_t, \textbf{A}_t, r, \textbf{O}_{t+T_r}, \textbf{C}'_{t+T_r}, \textbf{A}_{t+T_r})$ in demonstrations;
	(2) training the Guider according to Eq.~\ref{eq_32} and Eq.~\ref{eq_42};
	(3) training the Critic according to Eq.~\ref{eq_49};
	(4) training the Actor according to Eq.~\ref{eq_32}, Eq.~\ref{eq_45}, Eq.~\ref{eq_46}, and Eq.~\ref{eq_47}.
	At each time step during inference, the Guider first inputs the observation to predict the objective contact, and then the Actor inputs the observation and contact to predict the trajectory.

	\begin{figure}[tp]
		\centering
		{\includegraphics[scale=0.36]{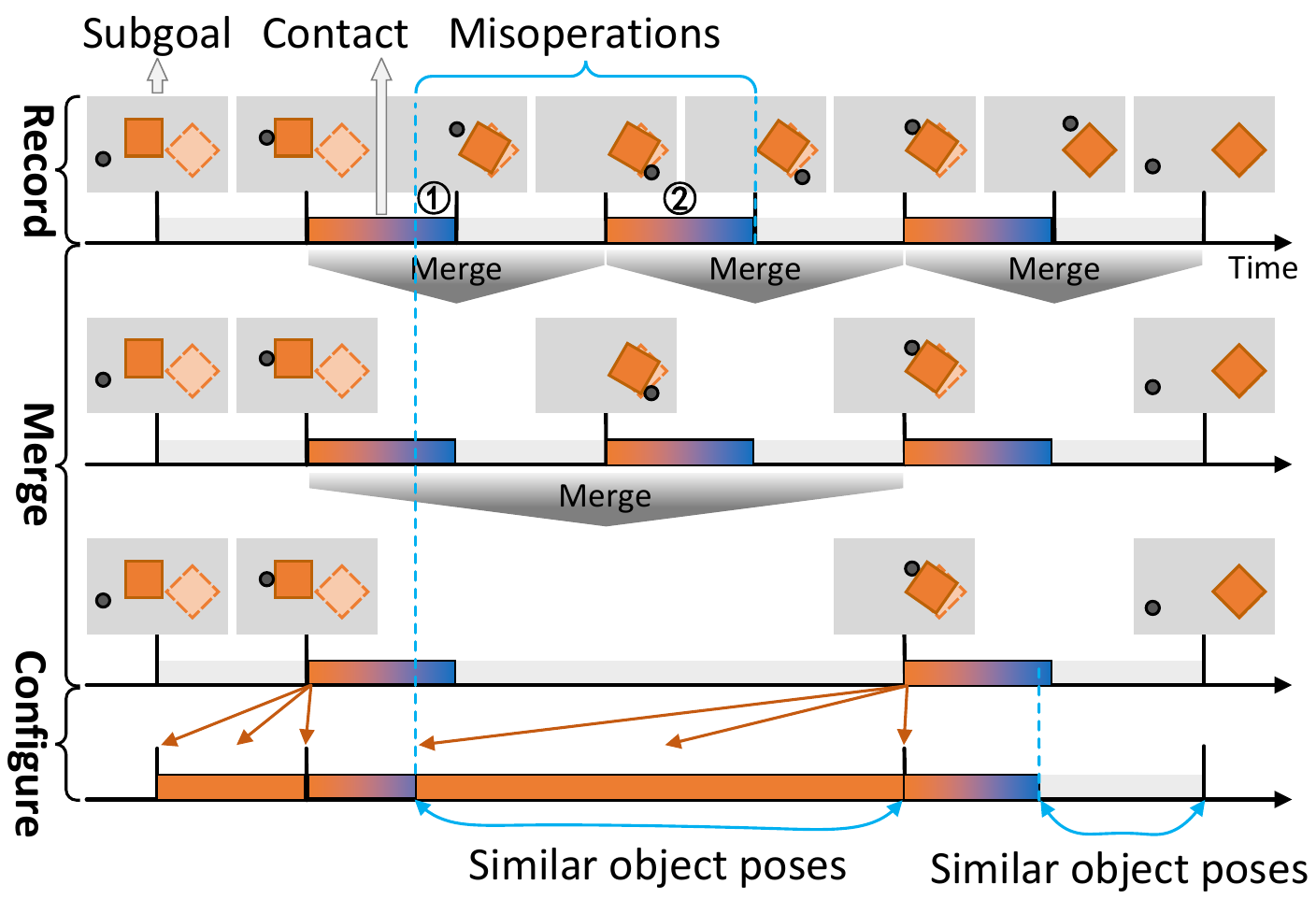}}
		\caption{
			\textbf{Phased objective contacts}.
			The algorithm consists of three steps: recording object subgoals and contacts, merging misoperations and no-contact operations, and configuring objective contacts.
			Misoperations include \textcircled{1} overshooting the movement and \textcircled{2} not moving or barely moving the object.
			The algorithm eliminates contacts related to misoperations by detecting object pose similarity, reducing suboptimal objective contacts (last row).
			Additionally, this process also removes the contacts related to operations near subgoals, resulting in an early update of the objective contact (the end of the last row), which is experimentally proven not to impair performance (Fig.~\ref{fig_result_oc}).
		}
		\label{fig_oc}
	\end{figure}

	\begin{algorithm}[tp]
		\caption{Phased objective contacts}
		\label{alg1}
		\KwIn{demonstration $D$ of length $N$,
			maximum reward $R$,
			Q-learning action steps $T_r$}
		\KwOut{objective contact $\textbf{C}_t$, $\textbf{C}'_t$ and reward $\textbf{R}_t$ for all $t \in [0, N-T_r]$}
		
		\SetKwFunction{FNA}{RecordFunction}
		\SetKwFunction{FNB}{MergeFunction}
		\SetKwFunction{FNC}{ConfigureFunction}
		
		// Step 1. Record the original contact status and object pose subgoal\\
		$\textbf{C}_o, \textbf{P}, \textbf{P}_I = $ \FNA{$D$};\\
		// Step 2. Merge contacts of no moving objects\\
		$\textbf{C}_o, \textbf{P}, \textbf{P}_I = $ \FNB{$\textbf{C}_o$, $\textbf{P}$, $\textbf{P}_I$};\\
		// Step 3. Configure objective contacts and rewards\\
		$\textbf{C}, \textbf{C}', \textbf{R} = $ \FNC{D, $\textbf{C}_o$, $\textbf{P}$, $\textbf{P}_I$, $R$};\\
		\Return{$\textbf{C}\leftarrow[\textbf{C}_0, ..., \textbf{C}_{N-T_r}]$, $\textbf{C}'\leftarrow[\textbf{C}'_0, ..., \textbf{C}'_{N-T_r}]$, $\textbf{R}\leftarrow[\textbf{R}_0, ..., \textbf{R}_{N-T_r}]$}
	\end{algorithm}
	
	\begin{algorithm}[tp]
		\caption{RecordFunction}
		\label{RecordFunction}
		\KwIn{demonstration $D$ of length $N$}
		\KwOut{original contact status $\textbf{C}_o$, 
			object pose subgoal $\textbf{P}$ and corresponding time steps $\textbf{P}_I$}
		
		$c_f \leftarrow false$; // Mark whether the robot is in contact with the object \\
		$\textbf{C}_o = \textbf{P} = \textbf{P}_I = \emptyset$;\\
		\For{$i=0$ to $N$}{
			update $c_f$, contact position $c_p$, object pose $p$;\\
			$\textbf{C}_o = \textbf{C}_o \cup (c_f, c_p)$;\\
			\If{$c_f$ changes}{
				$\textbf{P} = \textbf{P} \cup p$, $\textbf{P}_I = \textbf{P}_I \cup i$;\\
			}
		}
		\Return{$\textbf{C}_o$, $\textbf{P}$, $\textbf{P}_I$}
	\end{algorithm}
	
	\begin{algorithm}[tp]
		\caption{MergeFunction}
		\label{MergeFunction}
		\KwIn{original contact status $\textbf{C}_o$, 
			object pose subgoal $\textbf{P}$ and corresponding time steps $\textbf{P}_I$}
		\KwOut{merged contact status $\textbf{C}_o$, 
			object pose subgoal $\textbf{P}$ and corresponding time steps $\textbf{P}_I$}
		
		\While{$i<|\textbf{P}|$}{
			calculate similarity $\rho$ of $\textbf{P}[i]$ and $\textbf{P}[i+1]$ according to Eq.~\ref{eq_411};\\
			\eIf{$\rho > 0$}{
				\For{$j=\textbf{P}_I[i]$ to $\textbf{P}_I[i+1]$}{
					$\textbf{C}_o[j] \leftarrow (0, 0)$;
				}
				del $\textbf{P}[i], \textbf{P}_I[i]$;
			}{
				$i \leftarrow i+1$;
			}
		}
		
		del $\textbf{C}_o[0]$;\\
		$\textbf{C}_o = \textbf{C}_o \cup (0, 0)$; // Take the contact of the next time as the objective of the current time step\\
		\Return{$\textbf{C}_o$, $\textbf{P}$, $\textbf{P}_I$}
	\end{algorithm}
	
	\begin{algorithm}[tp]
		\caption{ConfigureFunction}
		\label{ConfigureFunction}
		\KwIn{demonstration $D$ of length $N$, 
			merged contact status $\textbf{C}_o$, 
			object pose subgoal $\textbf{P}$ and corresponding time steps $\textbf{P}_I$,
			maximum reward $R$, distance thresh $\tau_f$.}
		\KwOut{objective contact $\textbf{C}_t$, $\textbf{C}'_t$ and reward $\textbf{R}_t$ for all $t \in [0, N-T_r]$}
		
		$\textbf{C} = \textbf{C}' = \textbf{R} = \emptyset$;\\
		$p' \leftarrow 0$; // The serial number of the arrival object pose subgoal \\
		$d \leftarrow true$; // Mark whether the object reaches the subgoal \\
		$r \leftarrow 0$; // Reward \\
		\For{$i=0$ to $N-T_r$}{
			\If{$d = true$ and $r = R$}{
				$d = false$; // Update marker when robot reaches objective contact \\
			}
			
			\If{$d = false$}{
				calculate similarity $\rho$ of $p$ and $\textbf{P}[p'+1]$ according to Eq.~\ref{eq_411};\\
				\If{$\rho > 0$}{
					$p' \leftarrow p'+1$;\\
					$d = true$; // Update marker when object reaches the subgoal \\
				}
			}
			
			// Configure objective contact\\
			$i' = max(\textbf{P}_I[p'], i)$;\\
			$\textbf{C} = \textbf{C} \cup \textbf{C}_I[i']$;\\
			$c' \leftarrow$ objective contact at time $i+T_r$ and its position relative to the object is the same as $\textbf{C}_I[i']$;\\
			$\textbf{C}' = \textbf{C}' \cup c'$;\\
			
			// Calculate reward \\
			\For{$n=1$ to $T_r$}{
				$c \leftarrow$ objective contact at time $i+n$ and its position relative to the object is the same as $\textbf{C}_I[i']$;\\
				$f \leftarrow$ 	finger position;\\
				\eIf{$|f-c| < \tau_f$}{
					$r \leftarrow R$;\\
					break;
				}{
					$r \leftarrow 0$;
				}
			}
			$\textbf{R} = \textbf{R} \cup r$;\\	
		}
		
		\Return{$\textbf{C}, \textbf{C}', \textbf{R}$}
	\end{algorithm}

	\section{Key Design Decisions}
	In this section we outline the key design decisions for Hierarchical Diffusion Policy.
	
	\subsection{Objective Contact Position}
	\label{sec_oc}
	The first design decision is the choice of Guider's generation target, i.e., objective contact.
	The following misoperations usually occur in demonstrations:
	(1) the robot accidentally touches the object without moving it, and
	(2) does not stop properly after moving the object to the goal state, but instead corrects itself after the misoperation to complete the task.
	Taking the immediate contact as the objective would cause the policy to reproduce these misoperations during inference.

	
	We observe that the above misoperations have something in common, that is, they do not change the state of the object.
	Removing objective contacts that do not change the object's state may alleviate misoperations during inference.
	Based on this assumption, we design a novel method to construct phased objective contacts that exclude misoperations (Algorithm \ref{alg1}).
	
	The algorithm flow is illustrated in Fig.~\ref{fig_oc}.
	Specifically, the algorithm first divides the manipulation process into multiple phases according to the changes in contact state, and the object pose at the last time step of each phase is recorded as a subgoal (Algorithm \ref{RecordFunction}).
	The changed contact refers to switching from touching to detaching or vice versa, independent of changes in the contact position.
	Then the algorithm merges adjacent phases with similar subgoals and removes the subgoal from the previous phase and contacts within the next phase (Algorithm \ref{MergeFunction}).
	This process not only merges misoperations but also combines the phase where the object has never been touched with the previous phase.
	At the time step with contact, the objective contact is set to the position of the finger with contact.
	Once the object reaches the subgoal, the objective contact is set to the contact position of the first step in the next stage, even if the current stage has not ended due to an misoperation (Algorithm \ref{ConfigureFunction}).
	According to Eq.~\ref{eq_49}, the misoperation requires more time steps to reach the objective contact, resulting in a smaller Q-value.
	Therefore, optimizing the policy by maximizing the Q-value can avoid generating misoperations and improve efficiency.
	
	In Algorithm \ref{alg1}, object pose similarity $\rho$ is calculated according to the following formula:
	\begin{align}
		\rho = 
		\begin{cases} 
			1, & \mbox{if } dl < \tau_l \mbox{ and } dr < \tau_r \\
			0, & \mbox{else}
		\end{cases}
		\label{eq_411}
	\end{align}
	where $dl$ and $dr$ are the Euclidean distance and angle between two object poses respectively, $\tau_l$ and $\tau_r$ are the distance and angle thresholds respectively.

	\subsection{3D Encoder}
	\label{sec_3d_encoder}
	
	The 3D encoder maps the raw object point clouds into a latent embedding $\textbf{O}_t$ and is trained end-to-end with the HDP.
	For rigid objects, point clouds are derived from the model and transformed into the world coordinate system according to the pose.
	For deformable objects, point clouds are captured by stereo cameras or consist of keypoints.
	Point clouds in each timestep are concatenated and then encoded to form $\textbf{O}_t$.
	We modify a standard PointNet \cite{qi2017pointnet} and use it as the encoder.
	The modifications include: (1) removing input and feature transformations to make the network sensitive to object pose; (2) removing all normalization processing to make the network sensitive to object size.
	This is crucial because the object's pose and size directly influence how operations are carried out.
	
	The comparison of our new 3D encoder with PointNet and PointNet++ \cite{qi2017pointnet++} is shown in Tab.~\ref{tab_encoders}.
	Our 3D encoder outperforms others in both performance and inference speed.
	The performance of HDP using PointNet and PointNet++ with normalization is even worse than without using a 3D encoder, indicating that normalization not only prevents the network from accurately perceiving object information but also interferes with the network's processing of other observation.

	\begin{table}[tp]
		\footnotesize 
		\begin{center}
			\resizebox{\linewidth}{!}{
				\begin{tabular}{l|ccc}
					\toprule[1pt]
					
					\multirow{2}*{Encoder} & \multicolumn{2}{c}{Success rate}& \multirow{2}*{Proportion of inference time}  \\
					
					& Tilt(50) & Square-ph(50) & \\
					
					\midrule[0.7pt]
					
					No 3D encoder & 0.66/0.49 & 0.78/0.62 & 1 \\
					PointNet \cite{qi2017pointnet} & 0.02/0 & 0.06/0.02 & 1 \\
					PointNet w/o BN & 0.84/0.66 & 0.72/0.57 & 1 \\
					PointNet++ \cite{qi2017pointnet++} & 0/0 & 0/0 & 6 \\
					PointNet++ w/o BN & 0.72/0.60 & 0.68/0.59 & 6 \\
					
					\midrule[0.7pt]
					
					Our & \textbf{0.86/0.74} & \textbf{0.84/0.74} & 1 \\
					
					\bottomrule[1pt]
				\end{tabular}
			}
		\end{center}
		\captionsetup{justification=justified, singlelinecheck=false}
		\caption{\textbf{Comparison of 3D encoders}. We present success rates with different checkpoint selection methods in the format of (max performance) / (average of last 10 checkpoints), with each averaged across 3 training seeds and 50 different environment initial conditions (150 in total). 
			The details of the testing tasks are in Sec.~\ref{sec_tasks}. The numbers in brackets indicate the number of demonstrations used for training.}
		\label{tab_encoders}
	\end{table}

	\begin{figure*}[tp]
		\centering
		{\includegraphics[scale=0.55]{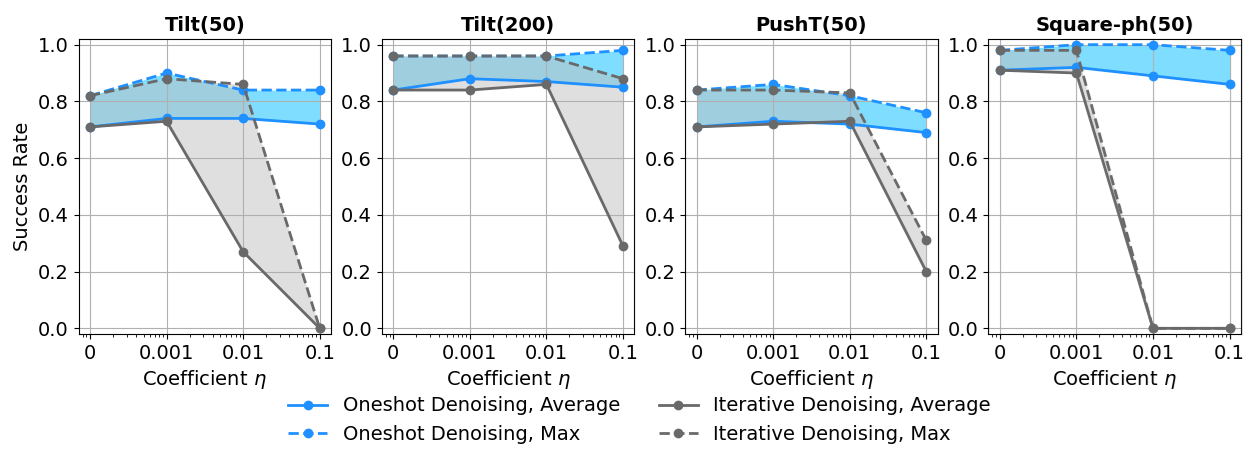}}
		\caption{
			\textbf{Q-learning Ablation Study}.
			HDP with oneshot denoising is better and robust than iterative denoising.
			A larger coefficient leads to greater optimization intensity and more optimization conflicts.
			The sensitivity of HDP to coefficient changes and gradient imbalance caused by iterative denoising increases as the number of training samples decreases.
		}
		\label{fig_eta_reverse}
	\end{figure*}

	\subsection{Data Augmentation}
	\label{data_augmentation}
	In the data tuples $(\textbf{O}_t, \textbf{C}_t, \textbf{A}_t, r, \textbf{O}_{t+T_r}, \textbf{C}'_{t+T_r}, \textbf{A}_{t+T_r})$ from the demonstration, $\textbf{A}_t$ is either suboptimal or optimal.
	The Critic trained on this data based on Eq.~\ref{eq_49} will only output similar Q values even for poor trajectories, thus failing to optimize the actor to generate the optimal trajectory.
	
	
	We alleviate this problem by adding noise to $\textbf{A}_t$ to expand the data distribution.
	Specifically, the added noise consists of two levels: minor and major.
	The minor noise slightly changes $\textbf{A}_t$ so that the observation after $T_r$ step is almost consistent with $\textbf{O}_{t+T_r}$.
	Conversely, major noise drastically changes the $\textbf{A}_t$ and the observation after $T_r$ step.
	The action after adding noise is formulated as:
	\begin{align}
		\tilde{\textbf{A}}_t = P (\textbf{A}_t + max(\tau_{n1}, & min(\tau_{n2}, \mathcal{N}(0, \mu_1\mathbf{I})))) 
		\notag
		\\ & + (1-P)(\textbf{A}_t + \mathcal{N}(0, \mu_2\mathbf{I}))
		\label{eq_431}
	\end{align}
	where $P$ takes $1$ or $0$ with the same probability, corresponding to minor and major noise respectively.
	$\mu_1$ and $\mu_2$ are used to adjust the noise variance.
	$\tau_{n1}$ and $\tau_{n2}$ are used to limit the amplitude of the minor noise, which does not exceed half the size of a finger in our experiments.
	The rewards corresponding to different noises are as follows:
	\begin{align}
		r = 
		\begin{cases} 
			r, & \mbox{if } P = 1 \\
			0, & \mbox{else}
		\end{cases}
		\label{eq_432}
	\end{align}
	We set the target Q-value of the trajectories with major noise to $0$ to suppress the motion patterns that significantly deviate from the demonstrations.
	Eq.~\ref{eq_49} is simplified to the following if $P=0$:
	\begin{align}
		\mathcal{L}_c = MSE(0, Q_{\phi}(\textbf{O}_t, \textbf{C}_t, \tilde{\textbf{A}}_t)
		\label{eq_433}
	\end{align}

	\subsection{Q-learning}
	\label{sec_q_learning}
	In the actor loss function established in Eq.~\ref{eq_45}, the design decisions related to Q-learning include the setting of the coefficient $\eta$ and the denoising formula of $\textbf{A}^0_{t,\theta}$.
	A comprehensive comparison of different coefficients $\eta$ and denoising formulas is shown in Fig.~\ref{fig_eta_reverse}.
	
	
	We find that HDP with the oneshot denoising formula consistently outperforms HDP with iterative denoising.
	This surprising result stands in contrast to the majority of recent diffusion reinforcement learning work that generally relies on iterative denoising \cite{wang2022diffusion, ada2024diffusion, black2023training}.
	We speculate that the main reason for this performance difference is that the network weights require different optimization intensities at different time steps.
	The data at the earlier inverse diffusion time steps contains more noise components, so the network needs to learn stronger denoising ability, resulting in higher optimization intensity and larger gradient.
	Iterative denoising does not balance the gradient of each step when jointly optimizing the weights, which weakens the optimization intensity at smaller time steps and causes the generated trajectory to deviate from the target distribution.
	In contrast, oneshot denoising ensures the best optimization intensity by individually optimizing the weights at each time step.
	Additionally, oneshot denoising increased the training speed by 4 times and reduced the GPU memory usage by 24\% when the denoising step is set to 10.
	We name the method of optimizing the Actor through oneshot denoising as snapshot gradient optimization.
	
	The experiment also verifies that a coefficient greater than 0 can help the policy reduce misoperations, but a coefficient that is too large will make the optimization intensity of Q-learning too high and conflict with imitation learning.
	We find that the coefficient of $0.001$ is optimal for most of the tasks we tested.
	With fewer demonstrations, the improvement from oneshot denoising is greater, and the policy is more sensitive to changes in the coefficient.

	\begin{figure}[tp]
		\centering
		{\includegraphics[scale=0.45]{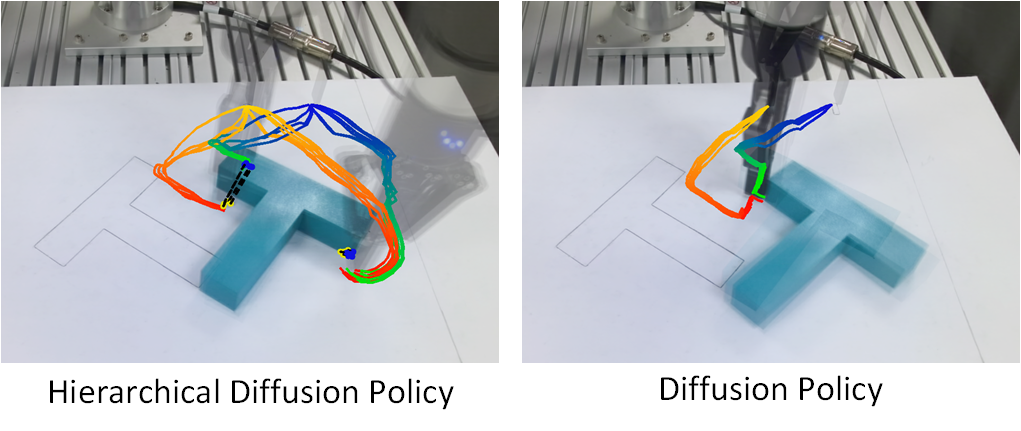}}
		\caption{
			\textbf{Multimodal behavior.}
			In the initial state, the robot can pull the left side or push the right side to move the object to the target position (black wireframe).
			The warm and cold curves represent the motion trajectories of the left and right fingers of the gripper, respectively.
			The two end points of the black dashed line are the objective contacts predicted by the Hierarchical Diffusion Policy.
			\textbf{Hierarchical Diffusion Policy} learns both patterns and executes exactly one of them in each deployment.
			\textbf{Diffusion Policy} is biased toward one mode, and has little success.
			Trajectories are generated by executing 10 times.
		}
		\label{fig_decompose}
	\end{figure}

	\begin{figure*}[tp]
		\centering
		{\includegraphics[scale=0.5]{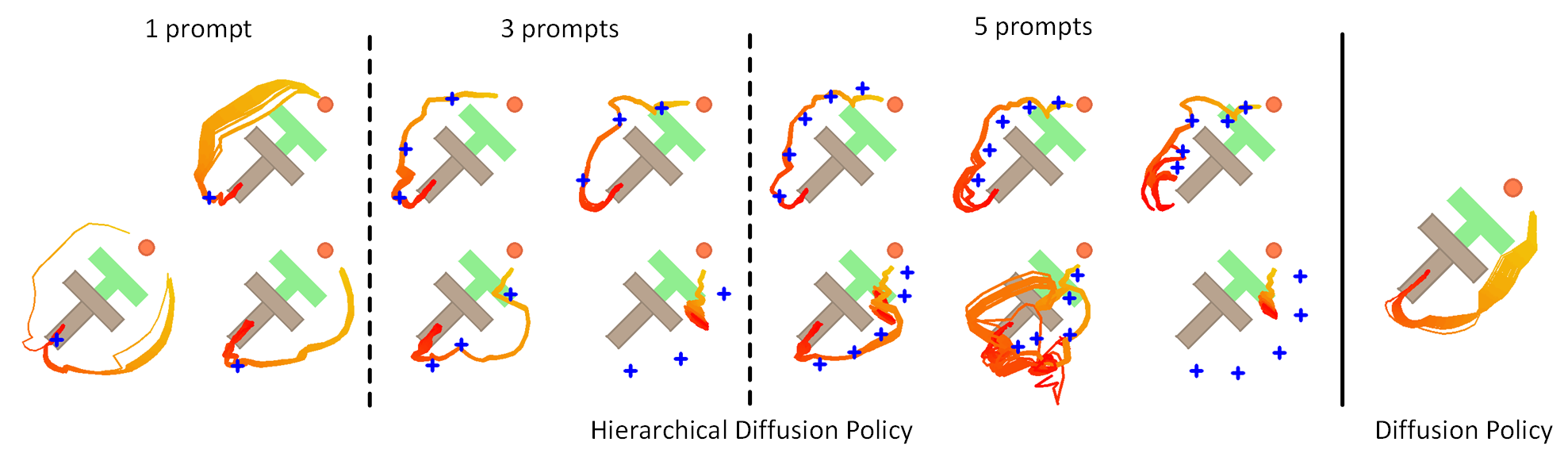}}
		\caption{
			\textbf{Prompt guidance}.
			At the given state, the end-effector (orange) can either go left or right to push the block.
			The blue cross is the manually designated target contact, i.e. prompt.
			\textbf{Hierarchical Diffusion Policy} can sequentially generate trajectories towards each prompt and perform the task.
			When the end-effector passes through a prompt, the input to the Actor switches to the next prompt.
			\textbf{Diffusion Policy} exhibits consistent movement patterns and could not be intervened by humans.
			Trajectories are generated by executing 50 times.
		}
		\label{fig_control}
	\end{figure*}

	\section{Intriguing Properties of HDP}
	In this section, we provide some insights and intuitions about hierarchical diffusion policy and its advantages over other forms of policy representations.
	
	\subsection{Decompose Multi-Modal Action Distributions}
	\label{sec_decompose}
	The challenges of modeling multi-modal distributions in demonstrations have been widely discussed in the literature \cite{florence2022implicit}\cite{ shafiullah2022behavior}\cite{mandlekar2022matters}.
	The key advantage of HDP is that it decomposes the problem of directly modeling multi-modal action distributions into two sub-problems: modeling the objective contact and the action with simpler distributions, which effectively reduces the modeling complexity and improves accuracy.
	
	In previous end-to-end modeling methods $f$, for a given observation $\textbf{O}$, the action distribution $\mathcal{A}$ typically includes multiple different contact strategies, expressed as follows:
	\begin{align}
		& f(\textbf{O}) = \textbf{A} \\
		\notag
		& \text{subject to} \quad \mathcal{C}_\mathcal{A} = \{\mathcal{C}_\textbf{A}|\textbf{A} \in \mathcal{A} \} = \{ \textbf{C}_1, \textbf{C}_2, \textbf{C}_3, ... \}
		\label{eq_511}
	\end{align}
	where $\mathcal{C}_\mathcal{A}$ is the set of contact strategies included in the action distribution $\mathcal{A}$.
	In contrast, due to the accurate guidance of Q-learning on the generated trajectories, the action distribution $\bar{\mathcal{A}}$ of HDP conditioned on the observation $\textbf{O}$ and objective contact $\textbf{C}$ only contains the given contact strategy:
	\begin{align}
		& f(\textbf{O}, \textbf{C}) = \bar{\textbf{A}} \\
		\notag
		& \text{subject to} \quad \mathcal{C}_{\bar{\mathcal{A}}} = \{\mathcal{C}_{\bar{\textbf{A}}}|\bar{\textbf{A}} \in {\bar{\mathcal{A}}} \} = \{\textbf{C}\}
		\label{eq_512}
	\end{align}
	Fewer contact strategies lead to a simpler multimodal distribution, reducing complexity and resulting in more stable predictions.
	Therefore, under the same policy modeling capabilities, HDP models the action distribution more accurately.
	Additionally, the powerful expressive capability of the diffusion model ensures the accuracy of modeling the objective contact $\textbf{C}$.
	Fig.~\ref{fig_decompose} shows an example of the HDP's multimodal behavior in a object reorientation task (Move-T, introduced below).
	Thanks to HDP's accurate objective contact prediction and trajectory guidance, the robot can successfully reach the appropriate operating position every time.

	\subsection{Prompt Guidance}
	With the rapid development of large language models, many methods use language models to guide the generation of robot motion trajectories, but most of them are limited to approximate spatial positions \cite{huang2023voxposer}\cite{driess2023palm}\cite{10549793}.
	Another key advantage of HDP is that humans can accurately guide the robot's trajectory by designating objective contacts even outside the training distribution.
	This ability stems from the fact that Q-learning suppresses trajectories that deviate from target contact.
	Thanks to the joint optimization of Q-learning and imitation learning, the trajectory does not go straight to the target contact, but remains close to the demonstrated behavior pattern.
	
	We name a manually designated objective contact as a prompt.
	Fig.~\ref{fig_control} shows an example of the HDP's prompt guidance in a planar pushing task (Push-T, introduced below).
	When prompts conform to the demonstrated trajectory distribution, HDP can generate effective operation trajectories that pass through each prompt in sequence.
	Although the objective contacts used for training are all located on the surface of objects, prompts outside the training distribution and far from the objects can still effectively guide the robot's motion.
	As the prompt gradually deviates from the demonstrated trajectory distribution, HDP experiences a conflict between following the prompt and performing the task as demonstrated.
	In contrast, Diffusion Policy \cite{chi2023diffusion} exhibits consistent movement patterns in the same test scenario and could not be intervened by humans.
	
	Prompt guidance enables the policy to cope with changes in task requirements without retraining, for example, constraining the range of motion of the end-effector in Push-T task.
	In addition, prompt guidance enhances HDP's potential to improve human-machine interaction, for example, correcting failed operations in rehearsed trajectories.

	\begin{figure}[tp]
		\centering
		{\includegraphics[scale=0.33]{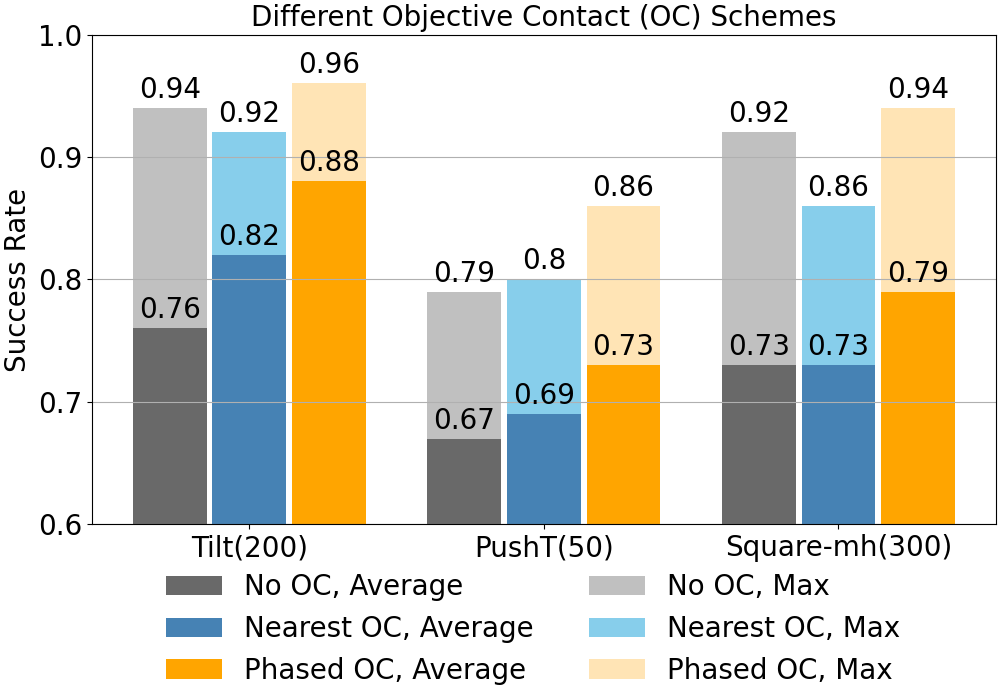}}
		\caption{
			\textbf{Different Objective Contact (OC) Schemes}.
			In the no-OC scheme, the Actor's condition is only observation.
			Compared with no-OC scheme, contact guidance improves the stability of the results.
			Compared with nearest-OC scheme, the phased-OC scheme improves the success rate by filtering out misoperations.
		}
		\label{fig_result_oc}
	\end{figure}

	\subsection{Benefits of Phased Contact}
	
	We test the performance of different objective contact schemes (Fig.~\ref{fig_result_oc}) in three tasks, and find two phenomena:
	(1) phased objective contact scheme has consistent advantages;
	(2) policies with objective contact guidance are more stable in results than policies without, that is, the difference between the average success rate and the maximum success rate is smaller.
	The nearest objective contact scheme takes the immediate contact as the objective.
	
	As described in Sec.~\ref{sec_oc} and Sec.~\ref{sec_q_learning}, the performance improvement of the phased objective contact scheme comes from filtering misoperations and accurate guidance.
	We attribute the improvement in stability to HDP being guided by contact, which aligns the generated trajectories more closely with the task objectives, rather than merely following the trajectory distribution learned during imitation learning. 
	This alignment enhances the consistency of the trajectories.
	The results in Fig.~\ref{fig_control} verify this conjecture, that is, when there is a prompt located to the right of the object, the consistency of HDP's trajectories is better than that of DP.

	\subsection{Effective for Any Q-Learning Action Steps}
\begin{figure}[tp]
		\centering
		{\includegraphics[scale=0.38]{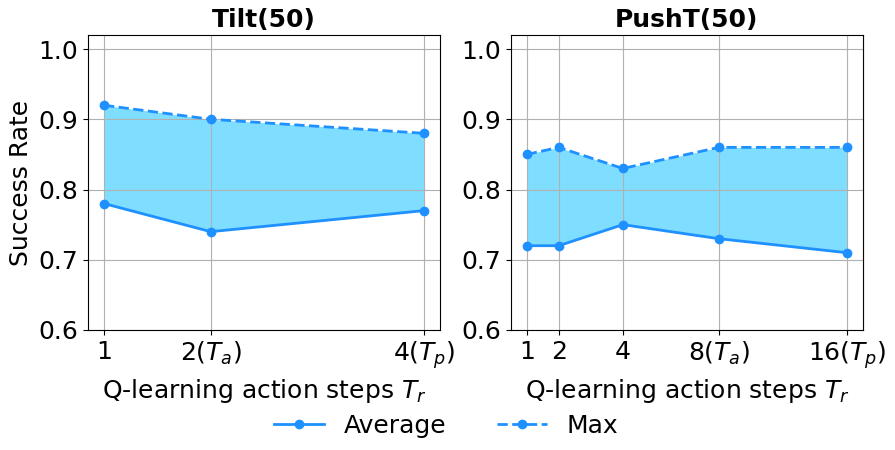}}
		\caption{
			\textbf{Q-learning Action Steps Ablation Study}.
			There is no significant difference in the success rate as $T_r$ increased.
		}
		\label{fig_T_r}
	\end{figure}

	We find that HDP is not sensitive to the number of Q-learning action steps (Fig.~\ref{fig_T_r}).
	This surprising result stands in contrast to the majority of recent reinforcement learning and trajectory generation work that is sensitive to the number of action steps \cite{kumar2023sample}\cite{wang2024multi}\cite{ze20243d}.
	We speculate that the reason for this discrepancy lies in the smoothness of the demonstration trajectories and the data continuity of the convolutional diffusion network.
	Even if Q-learning only optimizes the first $T_r$ actions, the subsequent actions generated by HDP (starting from step $T_r+1$) can still maintain continuity and smoothness.

	\begin{table*}[tp]
		\footnotesize 
		\begin{center}
			\resizebox{\linewidth}{!}{
				\begin{tabular}{l|ccc|ccc|ccc|ccc}
					\toprule[1pt]
					
					& \multicolumn{3}{c}{Can-ph} & \multicolumn{3}{c}{Can-mh} & \multicolumn{3}{c}{Square-ph} & \multicolumn{3}{c}{Square-mh}  \\
					
					& 30 & 100 & 200 & 30 & 100 & 300 & 50 & 100 & 200 & 50 & 100 & 300 \\
					
					\midrule[0.7pt]
					
					LSTM-GMM &-&-& \textbf{1.00}/0.91 &-&-& \textbf{1.00}/0.81 &-&-& 0.95/0.73 &-&-& 0.86/0.59 \\
					IBC &-&-& 0.00/0.00 &-&-& 0.01/0.01 &-&-& 0.00/0.00 &-&-& 0.00/0.00 \\
					BET &-&-& \textbf{1.00}/0.89 &-&-& \textbf{1.00}/0.90 &-&-& 0.76/0.52 &-&-& 0.68/0.43 \\
					
					Diffusion Policy & 0.88/0.81 & \textbf{1.00}/0.99 & \textbf{1.00}/0.97 & 0.82/0.73 & 0.96/\textbf{0.89} & \textbf{1.00}/0.95 & 0.78/0.62 & 0.92/0.83 & 0.98/0.89 & 0.40/0.28 & 0.72/0.61 & 0.90/0.76 \\
					
					\midrule[0.7pt]
					
					HDP (No-OC) & 0.92/\textbf{0.86} & \textbf{1.00}/0.99 & \textbf{1.00}/0.97 & \textbf{0.90}/\textbf{0.79} & 0.94/0.87 & \textbf{1.00}/0.95 & 0.84/\textbf{0.74} & \textbf{0.96}/\textbf{0.88} & 0.98/0.89 & \textbf{0.62}/\textbf{0.46} & 0.84/0.68 & 0.92/0.73 \\
					
					HDP ($\eta = 0$) & 0.90/0.81 & \textbf{1.00}/\textbf{1.00} & \textbf{1.00}/\textbf{0.99} & 0.82/0.73 & 0.94/0.85 & \textbf{1.00}/0.94 & \textbf{0.88}/0.72 & \textbf{0.96}/0.86 & 0.98/0.91 & \textbf{0.62}/0.44 & \textbf{0.86}/\textbf{0.71} & \textbf{0.96}/0.77 \\
					
					HDP ($\eta = 0.001$) & \textbf{0.94}/\textbf{0.86} & \textbf{1.00}/\textbf{1.00} & \textbf{1.00}/\textbf{0.99} & 0.82/0.74 & \textbf{0.98}/\textbf{0.89} & \textbf{1.00}/\textbf{0.97} & 0.74/0.60 & 0.94/0.86 & \textbf{1.00}/\textbf{0.92} & 0.56/0.44 & 0.82/0.69 & 0.94/\textbf{0.79} \\
					
					\midrule[0.7pt]
					
					HDP Improvement & 6.2\% & 1.0\% & 2.1\% & 8.2\% & 0\% & 2.1\% & 19.4\% & 6.0\% & 3.4\% & 64.3\% & 16.4\% & 3.9\% \\
					
					\bottomrule[1pt]
				\end{tabular}
			}
		\end{center}
		\captionsetup{justification=justified, singlelinecheck=false}
		\caption{\textbf{Comparison on Simulation Tasks (Part 1)}.
			Performance are reported in the same format as in Tab.~\ref{tab_encoders}.
			We reduce the size of the Diffusion Policy network and use the same backbone network in HDP to accelerate training, resulting in slightly lower performance compared to the results in \cite{chi2023diffusion}.
			We report the results of HDP and Diffusion Policy using different numbers of training samples to achieve a more detailed performance comparison.
			Our results show that Diffusion Policy significantly improves state-of-the-art performance across the board.
		}
		\label{tab_result_sim_1}
	\end{table*}

	\section{Evaluation}
	In this paper, we focus on testing the performance of HDP on tasks with single robot and single object.
	We systematically evaluate HDP on 6 tasks. 
	This evaluation suite includes both simulated and real environments, single and multiple task benchmarks, single and multiple operation modes, and rigid and deformable objects. 
	We find HDP to significantly outperform the prior state-of-the-art on all of the tested benchmarks, with an average success-rate improvement of 20.8\%. 
	In the following sections, we provide an overview of each task, our evaluation methodology on that task, and our key takeaways.

	\begin{figure}[tp]
		\centering
		{\includegraphics[scale=0.35]{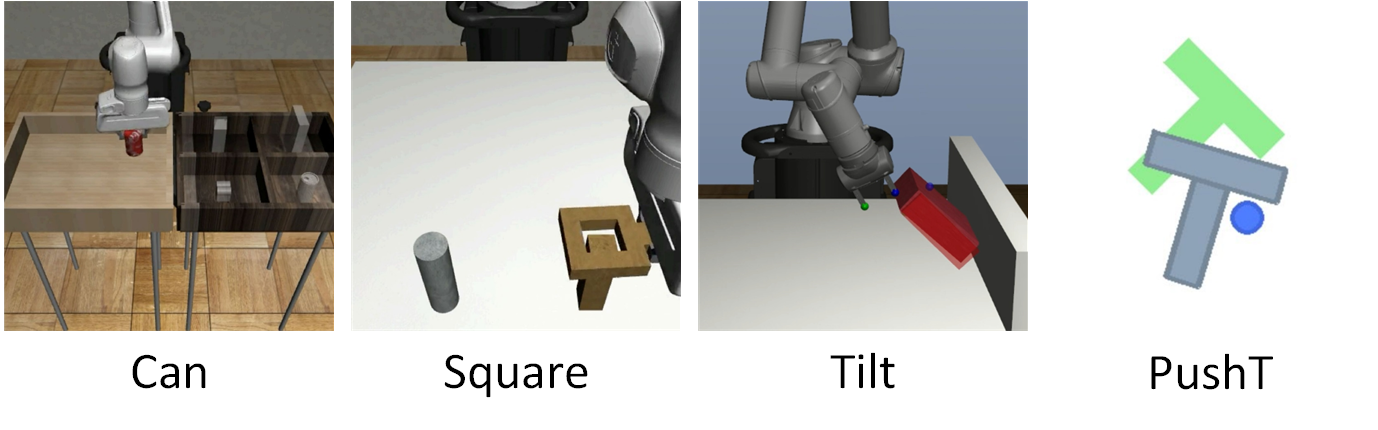}}
		\caption{
			\textbf{Simulation Tasks}.
			The Can and Square tasks require grasping, picking, and placing operations.
			The Tilt task requires pushing, tilting, grasping, and picking operations.
			The Push-T task requires pushing operations.
		}
		\label{fig_sim_tasks}
	\end{figure}

	\subsection{Simulation Environments and datasets}
	\label{sec_tasks}
	\textit{1) Robomimic}: \cite{mandlekar2022matters} is a large-scale robotic manipulation benchmark designed to study imitation learning and offline RL. 
	We choose 2 tasks in the benchmark with a proficient human (PH) teleoperated demonstration dataset and mixed proficient/non-proficient human (MH) demonstration datasets for each (4 variants in total). 
	The input observations include object pose and point cloud, and robot proprioceptive (end-effector pose and finger positions).
	Properties for each task are summarized in Tab.~\ref{tab_task_property}.

	\textit{2) Tilt}: adapted from \cite{wang2024multi}, requires pushing a cube that is too large to grasp from above to a distant wall, tilting it along the wall, and grasping it from the side.
	Variation is added by random initial position and size of the cube. 
	The task requires utilizing various operational skills and switching contact positions to accomplish precise quasi-static and non-quasi-static operations.
	The demonstrations are collected using the reinforcement learning method MRLNM \cite{wang2024multi} trained on this task.
	The input observations include object pose and point cloud, and robot proprioceptive (end-effector pose and finger positions).

	\textit{3) Push-T}: adapted from IBC \cite{florence2022implicit}, requires pushing a T-shaped block (gray) to a fixed target (green) with a circular end-effector (blue).
	Variation is added by random initial conditions for T block and end-effector. 
	The task requires exploiting complex and contact-rich object dynamics to push the T block precisely, using point contacts. 
	The input observations include 9 2D keypoints obtained from the ground-truth pose of the T block and proprioception for end-effector location.
	The visualization of the simulation tasks is shown in Fig.~\ref{fig_sim_tasks}.

	\subsection{Evaluation Methodology}
	For baseline methods, we compare with the latest diffusion-based end-to-end method (Diffusion Policy), sequence modeling-based methods (LSTM-GMM \cite{mandlekar2022matters}, BET \cite{shafiullah2022behavior}), and energy-based model (IBC).
	We reproduce the results of CNN-based Diffusion Policy with state-based input observations on all benchmarks due to inaccurate reported results \cite{li2023crossway}, and adopt the results of other methods reported in Ref.~\cite{chi2023diffusion}.
	We report results from the average of the last 10 checkpoints (saved every 50 epochs) across 3 training seeds and 50 environment initializations. 
	The metric for most tasks is success rate, except for the Push-T task, which uses target area coverage.
	The Actor networks in all tasks are trained for 3000 epochs, and the Guider and Critic networks are trained for 300k steps.
	The results from simulation benchmarks are summarized in Tab.~\ref{tab_result_sim_1} and Tab.~\ref{tab_result_sim_2}.
	The difference between HDP with no-OC and Diffusion Policy is that a 3D encoder is added.
	For the Push-T task, we concatenate the object keypoints with the proprioception and encode them together without using a 3D encoder, just like Diffusion Policy, to exploit the orderliness of the keypoints.
	The hyperparameters of the experiments are listed in the Appendix.

	\begin{table}[tp]
		\footnotesize 
		\begin{center}
			\resizebox{\linewidth}{!}{
				\begin{tabular}{l|ccc|ccc}
					\toprule[1pt]
					
					& \multicolumn{3}{c}{Tilt} & \multicolumn{3}{c}{Push-T} \\
					
					& 50 & 100 & 200 & 50 & 100 & 200 \\
					
					\midrule[0.7pt]
					
					LSTM-GMM &-&-&-&-&-& 0.67/0.61  \\
					IBC &-&-&-&-&-& 0.90/0.84  \\
					BET &-&-&-&-&-& 0.79/0.70  \\
					
					Diffusion Policy & 0.66/0.49 & 0.64/0.48 & 0.76/0.63 & 0.79/0.67 & 0.97/\textbf{0.87} & \textbf{1.00}/0.95 \\
					
					\midrule[0.7pt]
					
					HDP (No-OC) & 0.86/\textbf{0.74} & 0.86/0.76 & 0.94/0.76 & 0.79/0.67 & 0.97/\textbf{0.87} & \textbf{1.00}/0.95 \\
					
					HDP ($\eta = 0$) & 0.82/0.71 & 0.88/0.76 & \textbf{0.96}/0.84 & 0.84/0.71 & \textbf{0.98}/0.86 & \textbf{1.00}/\textbf{0.97} \\
					
					HDP ($\eta = 0.001$) & \textbf{0.90}/\textbf{0.74} & \textbf{0.96}/\textbf{0.82} & \textbf{0.96}/\textbf{0.88} & \textbf{0.86}/\textbf{0.73} & 0.94/0.83 & \textbf{1.00}/0.96 \\
					
					\midrule[0.7pt]
					
					HDP Improvement & 51.0\% & 70.8\% & 39.7\% & 9.0\% & 0\% & 2.1\% \\
					
					\bottomrule[1pt]
				\end{tabular}
			}
		\end{center}
		\captionsetup{justification=justified, singlelinecheck=false}
		\caption{\textbf{Comparison on Simulation Tasks (Part 2)}.
			Performance are reported in the same format as in Tab.~\ref{tab_result_sim_1}.
			For the Push-T task, HDP does not use the 3D encoder to encode key points, so the results of HDP with no-OC are the same as Diffusion Policy.
		}
		\label{tab_result_sim_2}
	\end{table}

	\begin{table}[tp]
		\footnotesize 
		\begin{center}
			\resizebox{\linewidth}{!}{
				\begin{tabular}{c|cccccccc}
					\toprule[1pt]
					
					Task & PH & MH & AP & Steps & PtN & ProD & ActD & HiPrec \\
					
					\midrule[0.7pt]
					\multicolumn{9}{c}{Simulation Benchmark} \\
					\midrule[0.7pt]
					
					Can & 200 & 300 & 0 & 400 & 1024 & 13 & 7 & No \\
					Square & 200 & 300 & 0 & 400 & 1024 & 13 & 7 & Yes \\
					Tilt & 0 & 0 & 200 & 200 & 1024 & 13 & 7 & Yes \\
					Push-T & 200 & 0 & 0 & 300 & 9 & 2 & 2 & Yes \\
					
					\midrule[0.7pt]
					\multicolumn{9}{c}{Realworld Benchmark} \\
					\midrule[0.7pt]
					
					Move-T & 192 & 0 & 0 & 800 & 700 & 13 & 5 & Yes \\
					Cloth Unfold & 100 & 0 & 0 & 300 & 4 & 13 & 7 & Yes \\
					
					\bottomrule[1pt]
				\end{tabular}
			}
		\end{center}
		\captionsetup{justification=justified, singlelinecheck=false}
		\caption{\textbf{Tasks Summary}.
			PH: proficient-human demonstration, MH: multi-human demonstration, AP: automated program demonstration, Steps: max number of rollout steps, PtN: number of object points, ProD: proprioceptive dimension, ActD: action dimension, HiPrec: whether the task has a high precision requirement.}
		\label{tab_task_property}
	\end{table}

	\subsection{Key Findings}
	HDP outperforms alternative methods on all tasks and variants in our simulation benchmark study (Tabs.~\ref{tab_result_sim_1} and \ref{tab_result_sim_2}) with an average improvement of 17.0\%.
	Following paragraphs summarize the key takeaways.
	
	\textbf{HDP can express short-horizon and  long-horizon multimodality}.
	Short-horizon action multimodality refers to the various ways of reaching the same objective contact.
	The leftmost case in Fig.~\ref{fig_control} shows the short-horizon multimodal behavior learned by HDP in the Push-T task, that is, approaching objective contact from the left and right sides respectively.
	Long-horizon action multimodality refers to the various ways of completing a task along different contact sequences.
	Fig.~\ref{fig_decompose} shows that HDP learns to complete the Move-T task with two different target contact sequences.
	Detailed experiments of HDP on the Move-T task are in Sec.~ \ref{sec_moveT}.

	\textbf{3D conditioning significantly improve performance}.
	In all tasks, HDP with no-OC achieves improvements.
	We find that the improvement is greater in the following three cases: (1) fewer training samples, (2) complex object shapes (e.g., Square), and (3) variable object shapes (e.g., Tilt).

	\textbf{HDP is better at learning optimal demonstrations.}
	In tasks requiring precise manipulation, the manually collected demonstration dataset contains many inadvertent touches that change the object's position which cannot be removed by Algorithm~\ref{alg1}, causing the Guider to learn objective contacts that reproduce these mistakes (e.g., Square task).
	In contrast, the demonstrations for the Tilt task collected by the RL policy are smoother and free of inadvertent touches.
	The results indicate that contact guidance provides an average improvement of 7.9\% on the Tilt task, but an average decrease of 2.1\% on the Square task (comparing HDP with $\eta=0.001$ to no-OC), with the most significant decrease occurring in low-sample scenarios.

	\section{Realworld Evaluation}
	
	
	We evaluate the performance of HDP on two real-world tasks, focusing on multimodal manipulation of rigid and deformable objects, respectively.
	In all comparisons, HDP outperforms Diffusion Policy, with an average improvement of 55.4\%.

	\begin{figure}[tp]
		\centering
		{\includegraphics[scale=0.5]{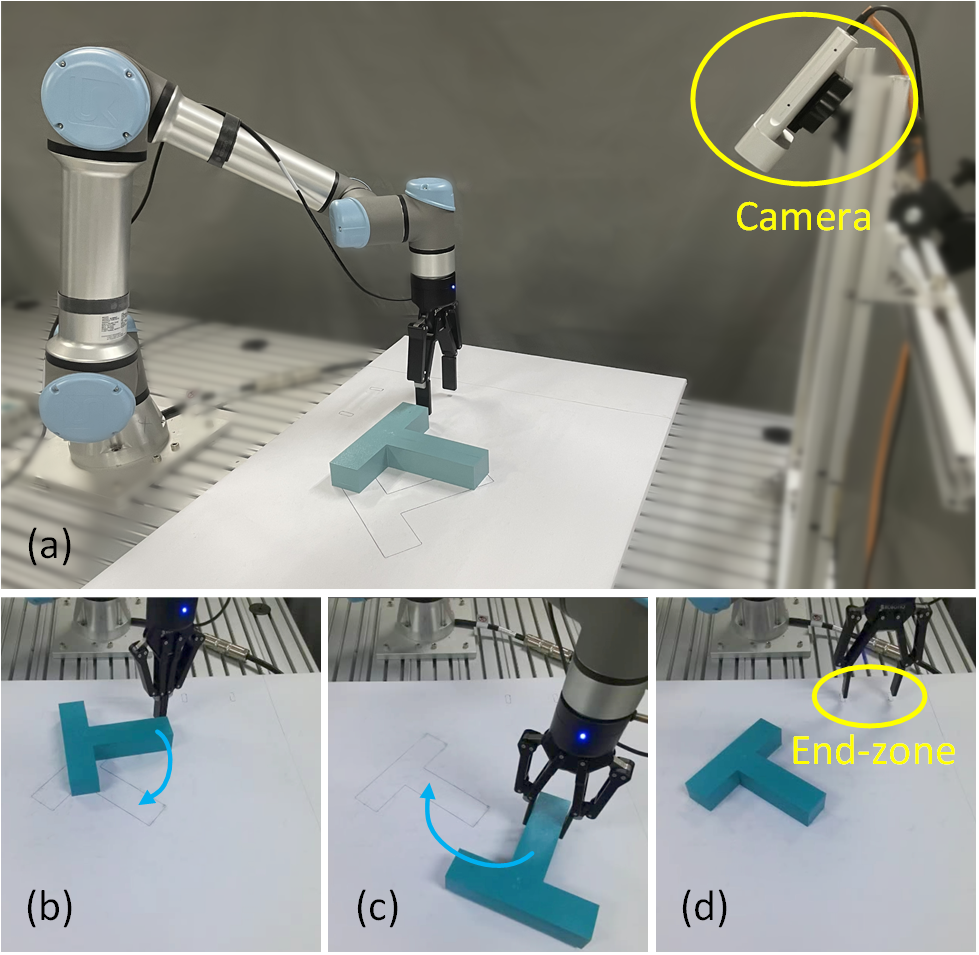}}
		\caption{
			\textbf{Realworld Move-T Experiment Setup}.
			(a) Hardware setup.
			The robot needs to precisely (b) push or (c) clamp the T-shaped block to the target region, and (d) then move the end-effector to the end-zone.
		}
		\label{fig_moveT_setup}
	\end{figure}

	\subsection{Move-T Task}
	\label{sec_moveT}
	
	The realworld Move-T task is an extension of the simulated Push-T task. The extensions mainly include the following:
	(1) the end-effector is a commonly used parallel gripper instead of a cylinder;
	(2) the operation modes include both push and grasp to fully utilize the gripper's capabilities;
	(3) the task is multi-stage, and the robot needs to first move the T block to the target position and then move the gripper into a designated end-zone;
	(4) the IoU metric is measured at the last step instead of taking the maximum over all steps.
	Our UR5e-based experiment setup is shown in Fig.~\ref{fig_moveT_setup}. 
	Each demonstration in the dataset contains only one operation mode of pushing or grasping.
	Action dimensions include 3D translation, rotation along the vertical axis, and manipulator closure.
	The input observations are object poses and point clouds as well as the robot's proprioception.
	The point cloud is generated by transforming the pre-constructed T block point cloud to the object positions obtained through Iterative Closest Point (ICP) alignment.
	For the Move-T task, Algorithm~\ref{RecordFunction} of HDP additionally records the gripper finger positions in the last step of the demonstration as objective contacts to provide guidance for the operation in stage 2.
	The action sequence predicted by HDP runs at 5 Hz.
	
	\begin{table}[tp]
		\footnotesize 
		\begin{center}
			\resizebox{\linewidth}{!}{
				\begin{tabular}{l|ccc|ccc}
					\toprule[1pt]
					
					& \multicolumn{3}{c}{Diffusion Policy} & \multicolumn{3}{c}{HDP}  \\
					
					& Push & Grasp & Both & Push & Grasp & Both \\
					
					\midrule[0.7pt]
					
					Succ. \% (Stage 1) & 0.20 & 0.00 & 0.15 & 0.15 & 0.00 & 0.05 \\
					Succ. \% (Stage 2) & 0.35 & 0.45 & 0.30 & \textbf{0.65} & \textbf{0.75} & \textbf{0.60}  \\
					
					\midrule[0.7pt]
					
					Aver. IoU (Stage 2) & \textbf{0.96} & 0.94 & \textbf{0.96} & 0.94 & \textbf{0.95} & 0.93 \\
					Aver. IoU           & 0.57 & 0.54 & 0.52 & \textbf{0.78} & \textbf{0.72} & \textbf{0.63} \\
					
					\bottomrule[1pt]
				\end{tabular}
			}
		\end{center}
		\captionsetup{justification=justified, singlelinecheck=false}
		\caption{\textbf{Results of Realworld Move-T Experiment}.
			All demonstrations are categorized into three datasets based on the operation mode: push, grasp, and a combination of both. 
			The datasets contain 128, 64, and 192 demonstrations, respectively.
			We report the results of 20 test runs for each method trained on the three datasets.
			At the end of the run, if the IoU exceeds the threshold (0.8 in our experiment) and the end-effector reaches the end-zone, stage 2 is considered successful; if only the IoU criterion is met, stage 1 is considered successful.
		}
		\label{tab_moveT_result}
	\end{table}

	\begin{table}[tp]
		\footnotesize 
		\begin{center}
			\resizebox{\linewidth}{!}{
				\begin{tabular}{l|ccc}
					\toprule[1pt]
					
					& Diffusion Policy & HDP & HDP w/ Prompt Guidance \\
					
					\midrule[0.7pt]
					
					Succ. \% (Stage 2) & 0.29 & 0.29 & \textbf{0.71} \\
					
					Aver. IoU          & 0.43 & 0.63 & \textbf{0.74} \\
					
					\bottomrule[1pt]
				\end{tabular}
			}
		\end{center}
		\captionsetup{justification=justified, singlelinecheck=false}
		\caption{\textbf{Prompt Guidance Experiment Results in Move-T Task.}
			Results are obtained on 7 test scenarios using policies trained on the push mode dataset.
			Prompt Guidance significantly improves the success rate and IoU.
		}
		\label{tab_moveT_prompt}
	\end{table}

	\begin{figure*}[tp]
		\centering
		{\includegraphics[scale=0.5]{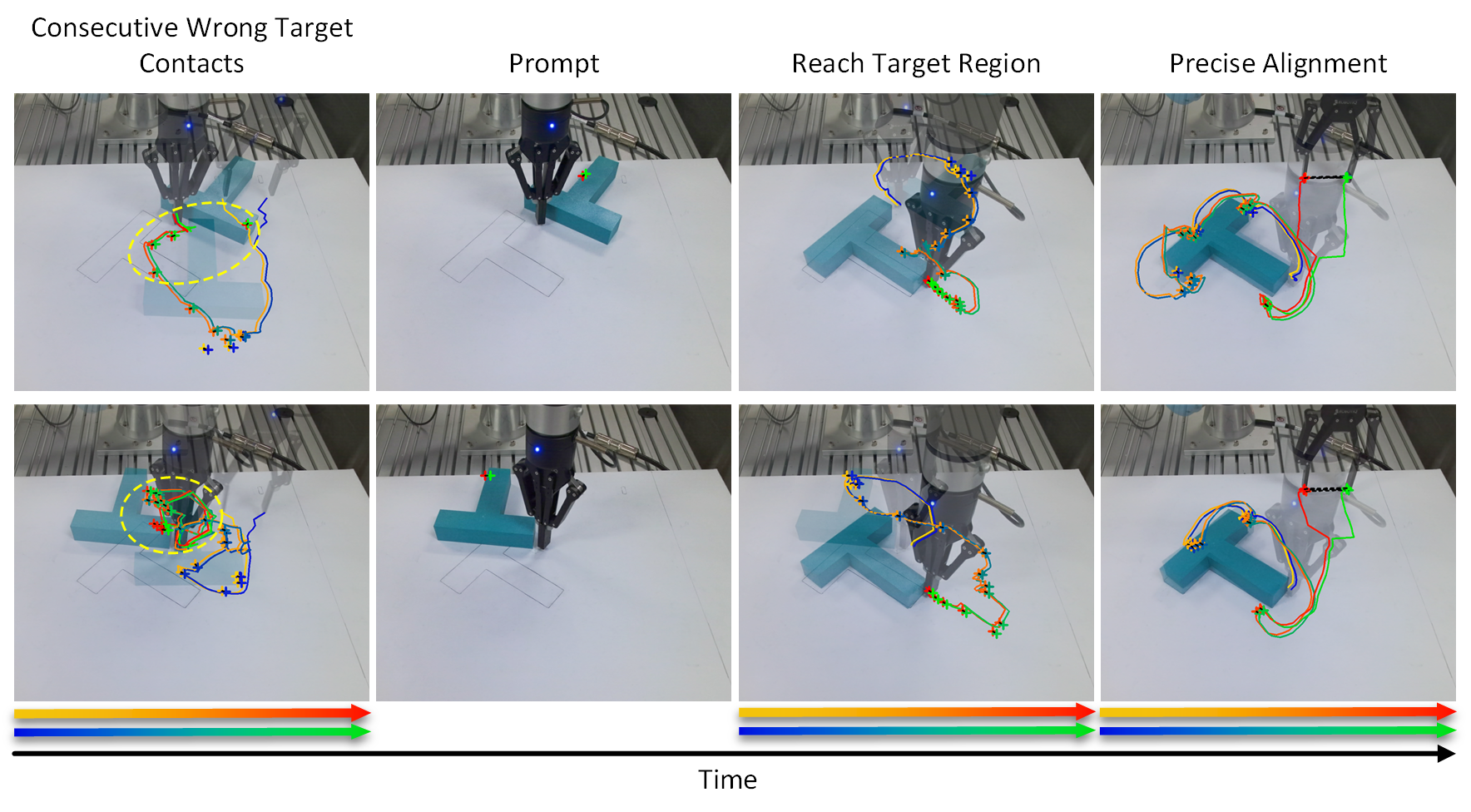}}
		\caption{
			\textbf{Prompt Guidance in Move-T Task}.
			Each row represents an execution example.
			When the robot is about to push the object off the table or fall into trajectory oscillation (within the yellow circle), a prompt can guide the robot to complete the task.
			The crosses in the second column are prompts, and the crosses in other figures are objective contacts predicted by the Guider.
		}
		\label{fig_moveT_prompt}
	\end{figure*}

	\begin{figure*}[tp]
		\centering
		{\includegraphics[scale=0.35]{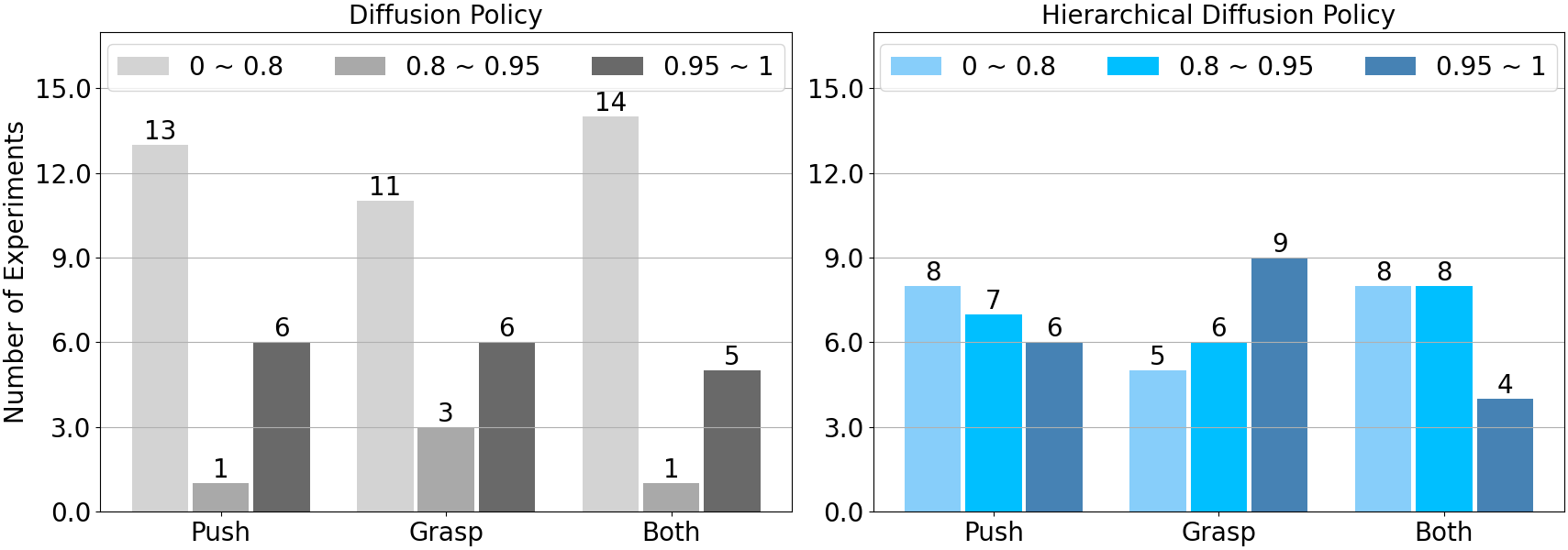}}
		\caption{
			\textbf{IoU Distribution of Realworld Move-T Experiment}.
			The IoU distribution of HDP is more balanced.
			The IoU of Diffusion Policy tends to be at either extreme, indicating that Diffusion Policy either completely fails or performs almost perfectly.
		}
		\label{fig_moveT_result}
	\end{figure*}

	\begin{figure}[tp]
		\centering
		{\includegraphics[scale=0.49]{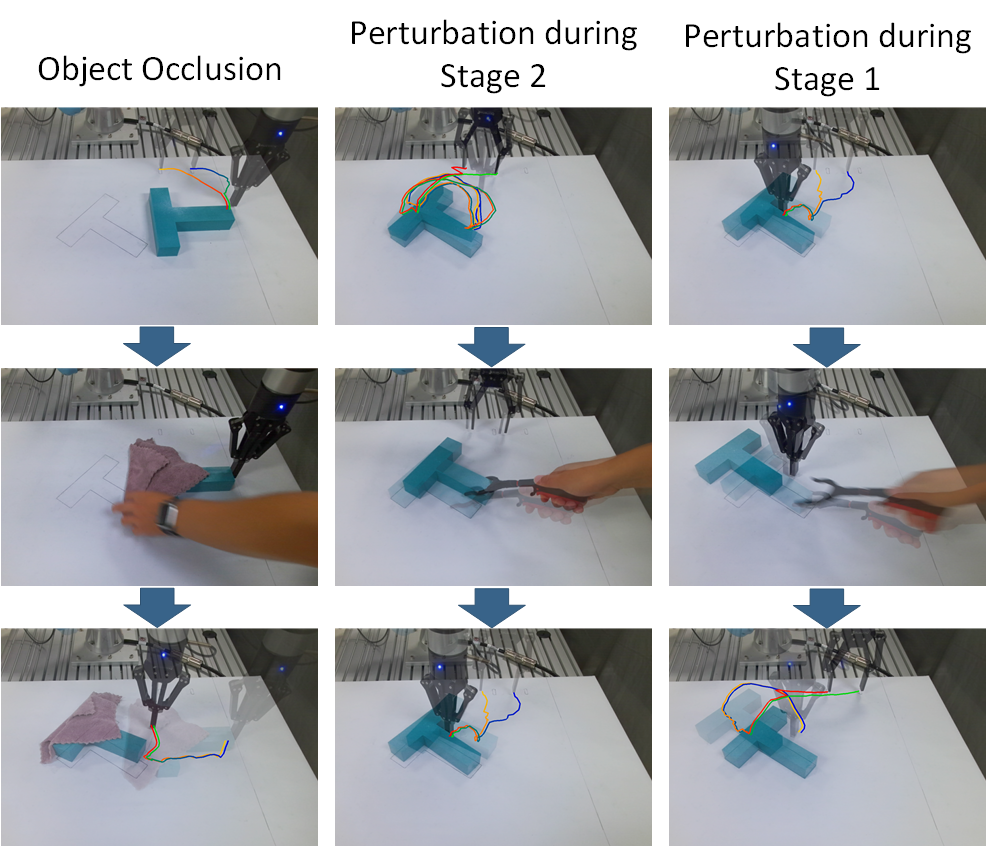}}
		\caption{
			\textbf{Robustness Test for hierarchical Diffusion Policy.}
			\textbf{Left}: The Block is blocked by a towel and can still be moved to the target area.
			\textbf{Middle}: HDP immediately aborts heading to the end-zone, returning the block to goal state upon detecting block shift.	
			\textbf{Right}: HDP immediately corrects the shifted block position to the target state.
		}
		\label{fig_moveT_perturbation}
	\end{figure}

	\textbf{Result Analysis}.
	The success rate and IoU are shown in Tab.~\ref{tab_moveT_result}.
	HDP improves the average success rate by 84.1\% compared to Diffusion Policy.
	From the comparison of the success rates, it can be concluded that the improvement of HDP mainly comes from two aspects:
	(1) a higher success rate in stage 2 indicates that HDP has a stronger ability to express multi-modal action distribution,
	(2) a lower success rate in stage 1 indicates that HDP is more effective in guiding the end-effector to move to the end-zone after the object reaches the target area.
	The improvement in expressiveness comes from the decomposition of complex multimodal distributions, which is analyzed in detail in Sec.~\ref{sec_decompose}.
	The minimum IoU in the demonstration dataset is 0.93. 
	This high termination condition causes the Diffusion Policy to continue executing stage 1 when the object's IoU is less than the demonstration threshold, preventing it from progressing to stage 2.
	In contrast, when the error between the object and the target ground truth is below the thresholds $\tau_l$ and $\tau_r$, Algorithm \ref{ConfigureFunction} of HDP configures the objective contact as the finger position within the end-zone. 
	Therefore, the Guider can predict the objective contact within the end-zone as the object's IoU approaches the demonstration value, guiding the robot to execute stage 2.
	This is demonstrated by the lower average IoU of stage 2 of HDP in Tab.~\ref{tab_moveT_result}.
	
	The detailed distribution of IoU results is shown in Fig.~\ref{fig_moveT_result}.
	The number of experiments with IoU exceeding 0.95 for both methods is similar, while the number of experiments with IoU between 0.8 and 0.95 for HDP is 320\% higher than that for Diffusion Policy, further proving that HDP can better learn multi-stage tasks.

	
	\textbf{Prompt guidance.}
	Quantitative results show that prompt guidance increases the success rate by 145\% (Tab.~\ref{tab_moveT_prompt}).
	In each test scenario, when the robot is about to push the block out of the table or exhibits trajectory oscillates due to consecutive wrong objective contacts predicted by the Guider, the supervisor will specify a objective contact through the visual interface to replace the one predicted by the Guider until the Guider predicts a suitable objective contact.
	In all test scenarios, the number of prompts given by the supervisor does not exceed 3.
	Two examples of prompt guidance are shown in Fig.~\ref{fig_moveT_prompt}.
	Humans can guide the robot out of a local optimum trap or trajectory oscillation using prompts.
	This type of human-robot interaction has never been seen in previous imitation learning methods.

	\textbf{Robustness against perturbation.}
	We test the robustness of HDP against three types of perturbation in an uninterrupted run:
	(1) being blocked by an object, (2) moving the block during the robot's return to the end zone, and (3) moving the object during robot operation.
	As shown in Fig.~\ref{fig_moveT_perturbation}, the robot reacts quickly and continues to move the object to the target area without exhibiting abnormal behavior.
	None of these perturbations appear in the demonstration, indicating that HDP can generalize effectively to unseen observations.
	Please check our website for the full test video.

	\begin{figure}[tp]
		\centering
		{\includegraphics[scale=0.55]{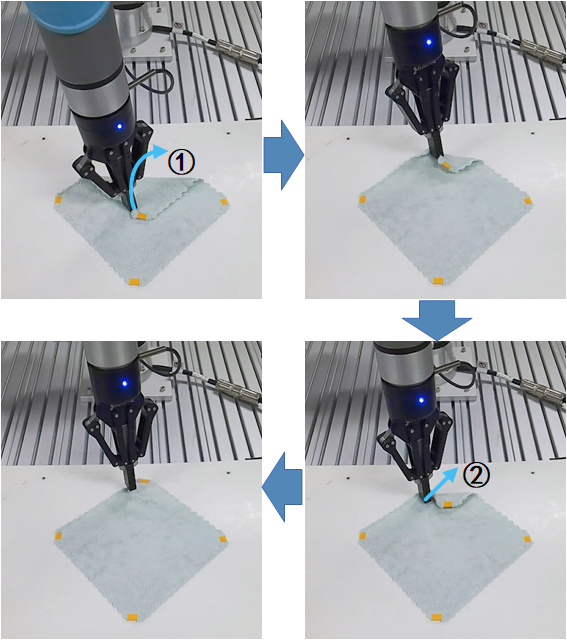}}
		\caption{
			\textbf{Cloth Unfolding Experiment Setup}.
			The robot needs to use either \textcircled{1} picking or \textcircled{2} pushing to unfold the folded cloth, and sometimes a combination of these two modes of operation is required.
			The yellow markers are used to detect the four corner positions.
		}
		\label{fig_cloth_setup}
	\end{figure}

	\subsection{Cloth Unfolding Task}
	We design the cloth unfolding task to test the performance of HDP on deformable objects.
	The task objective is to use a parallel gripper to unfold a square piece of cloth with one corner folded.
	The challenges include the combined use of multiple operation modes, a compact operation space, and a 7-dimensional action space.
	(1) When the folded part is large, the robot needs to clamp the folded edge to unfold it.
	When the folded part is small, direct pushing works better.
	During operation, clamping and pushing sometimes need to be used in combination because the cloth may not be unfolded in one go.
	(2) For folded cloth, the positions available for clamping and pushing are limited to the edges, with the height range not exceeding the thickness of the cloth.
	If two layers of cloth are mistakenly lifted, it will make the task more difficult to handle.
	(3) The high operation dimension increases the difficulty of learning.
	Our UR5e-based experiment setup is shown in Fig.~\ref{fig_cloth_setup}. 
	The input observations are the robot proprioception and the positions of the four corner points of the cloth, which is obtained by detecting markers of specific colors.
	
\begin{figure*}[tp]
		\centering
		{\includegraphics[scale=0.33]{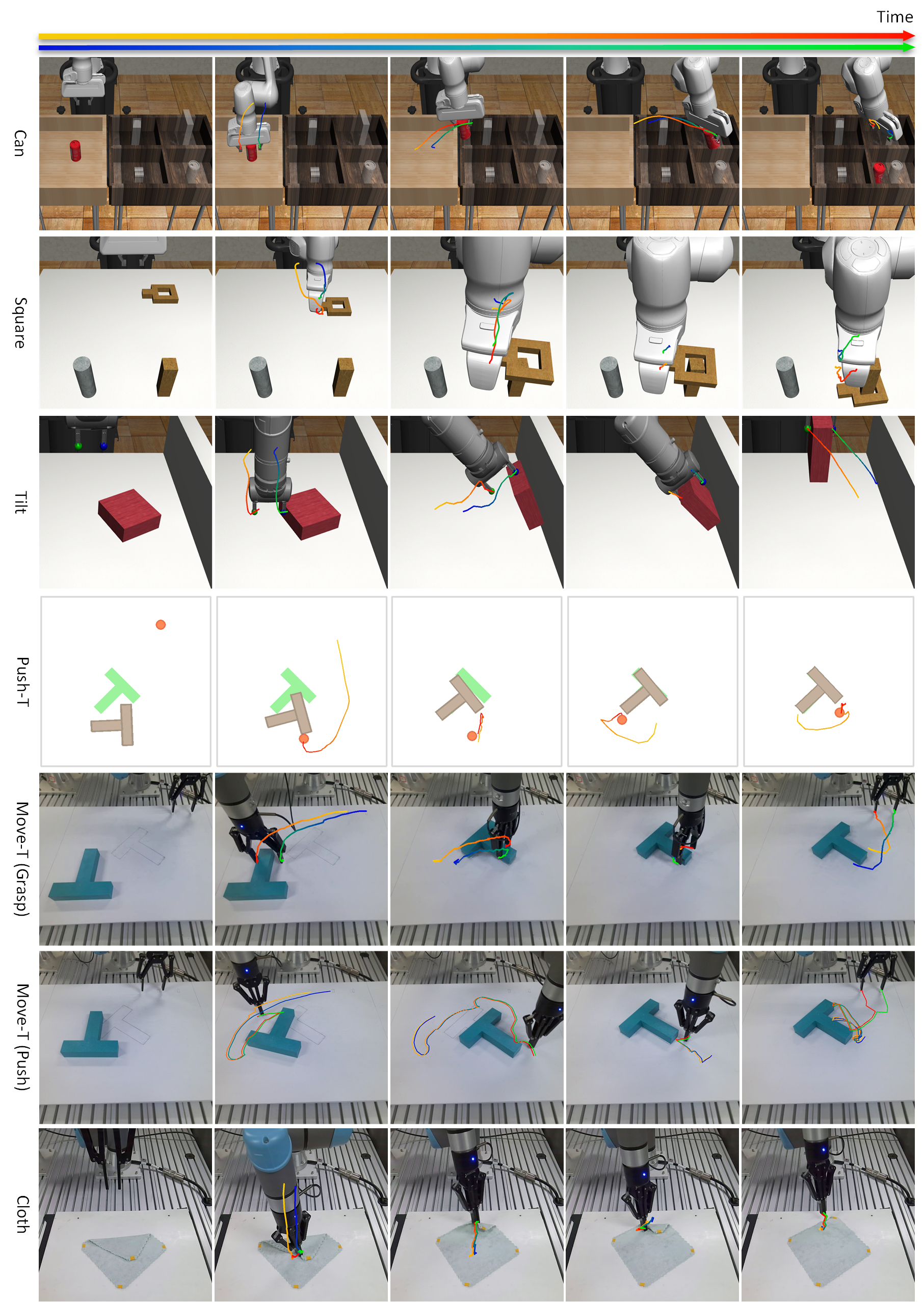}}
		\caption{
			\textbf{Example of each test task}.
			For Move-T task, we show running examples of two policies trained on demonstrations containing only pushing or grasping operation modes, respectively.
		}
		\label{fig_result_all}
	\end{figure*}

	\textbf{Result Analysis}.
	HDP achieves a success rate of 76\%, which is 26.7\% higher than Diffusion Policy.
	Although never demonstrated, HDP attempts repeated clamping and lifting in the event of a lifting failure until successful; the same applies to push failures.
	The clamping position of Diffusion Policy is sometimes low, causing both layers of cloth to be lifted.
	In contrast, HDP is better at clamping the correct position.
	A unique failure case produced by HDP occurs when lifting causes other previously unfolded corners to fold up. 
	This happens because the clamping position is too close to the middle of the cloth's edge, causing the other corners to be pulled up during lifting.
	The running example of each test task is shown in Fig.~\ref{fig_result_all}.

	\begin{table}[tp]
		\footnotesize 
		\begin{center}
			\resizebox{\linewidth}{!}{
				\begin{tabular}{l|ccc|c}
					\toprule[1pt]
					
					& \multicolumn{3}{c|}{Failure Count} & \multirow{2}*{Succ \%}  \\
					
					& Planning Failure & Lift Both Layers & Other Corners Folded &  \\
					
					\midrule[0.7pt]
					
					Diffusion Policy & 6 & 14 & 0 & 0.60  \\
					HDP & 6 & 2 & 4 & \textbf{0.76}  \\
					
					\bottomrule[1pt]
				\end{tabular}
			}
		\end{center}
		\captionsetup{justification=justified, singlelinecheck=false}
		\caption{\textbf{Results of Cloth Unfolding Experiment}.
			Each method is tested 50 times.
			Planning failures include situations where the robot cannot lift or push the cloth.
		}
		\label{tab_cloth_result}
	\end{table}

	\begin{table*}[tp]
		\footnotesize 
		\begin{center}
			\resizebox{\linewidth}{!}{
				\begin{tabular}{l|ccccccccccccc}
					\toprule[1pt]
					
					Task & Ctrl & $T_o$ & $T_p$ & $T_a$ & $T_r$ & $\tau_l$ & $\tau_r$ ($^\circ$) & $\tau_f$ & SEef & $R$ & DimOC & RepPred & DimPred  \\
					
					\midrule[0.7pt]
					
					Can 	& Pose 	& 2 & 16 & 8 & 8 & 0.02 m & 10 & 0.004 m & 0.016 m & 10 & 8 & Pos+6DRot+Close & 10 \\
					Square 	& Pose 	& 2 & 16 & 8 & 8 & 0.02 m & 10 & 0.004 m & 0.016 m & 10 & 8 & Pos+6DRot+Close & 10 \\
					Tilt 	& Pose 	& 2 & 4  & 2 & 2 & 0.02 m & 10 & 0.004 m & 0.016 m & 10 & 8 & Pos+AA+Close & 7 \\
					Push-T 	& Pos 	& 2 & 16 & 8 & 8 & 3 	  & 1  & 7.5 & 30 & 10 & 3 & Pos & 2 \\
					
					\midrule[0.7pt]
					
					Move-T 			& Pose & 2 & 16 & 8 & 8 & 0.002 m & 1 & 0.0075 m & 0.03 m & 10 & 8 & Pos+6DRot+Close & 10 \\
					Cloth Unfold 	& Pose & 2 & 16 & 8 & 8 & - & - & - & - & 10 & 8 & Pos+6DRot+Close & 10 \\
					
					\bottomrule[1pt]
				\end{tabular}
			}
		\end{center}
		\captionsetup{justification=justified, singlelinecheck=false}
		\caption{\textbf{Hyperparameters for Tasks}.
			Ctrl: control mode of the robot end-effector, the pose includes position and rotation,
			$T_o$: observation horizon,
			$T_p$: action prediction horizon,
			$T_a$: action execution horizon,
			$T_r$: Q-learning action steps,
			$\tau_l$: distance threshold for assessing object similarity,
			$\tau_r$: angle threshold for assessing object similarity,
			$\tau_f$: distance threshold for reaching the objective contact,
			SEef: cross-sectional size of the end-effector,
			$R$: maximum reward,
			DimOC: dimensions of objective contact,
			RepPred: representation of predicted actions, rotation representation is 6D rotation \cite{zhou2019continuity} or axis-angle,
			DimPred: dimensions of predicted actions.
		}
		\label{tab_hyperparameters_task}
	\end{table*}

	\begin{table*}[tp]
		\footnotesize 
		\begin{center}
			\resizebox{\linewidth}{!}{
				\begin{tabular}{l|cccccccccccc}
					\toprule[1pt]
					
					Network & Batch & Epoch & Steps & Lr & WDecay & $\eta$ & $\gamma$ & Params & DIters & $\sigma$ & $\mu_1$ & $\mu_2$ \\
					
					\midrule[0.7pt]
					
					Actor & 256 & 3000 & - & 1e-4 & 1e-6 & 1e-3 & - & 9.5 $\pm$ 0.2 & 10 & 0.75 & - & -  \\
					Guider & 256 & - & 3e5 & 1e-3 & 1e-6 & - & - & 1.2 $\pm$ 0.1 & 100 & 0.75 & - & -  \\
					Critic & 256 & - & 3e5 & 1e-3 & 1e-6 & - & 0.95 & 0.8 $\pm$ 0.4 & - & 0.005 & 0.1 & 1  \\
					
					\bottomrule[1pt]
				\end{tabular}
			}
		\end{center}
		\captionsetup{justification=justified, singlelinecheck=false}
		\caption{\textbf{Hyperparameters for Networks}.
			Steps: number of training iteration steps,
			Lr: learning rate,
			WDecay: weight decay,
			$\eta$ and $\gamma$: loss function coefficient,
			Params: number of network parameters in millions,
			DIters:  number of diffusion iterations,
			$\sigma$: moving average coefficient,
			$\mu_1$ and $\mu_2$: Data enhancement coefficient.
		}
		\label{tab_hyperparameters_net}
	\end{table*}

	\section{Discussion}
	Although HDP has demonstrated effective learning and generalization capabilities in the selected tasks, it still has some limitations and room for improvement.
	Firstly, the prompt guidance experiment in Fig.~\ref{fig_control} on the Push-T task shows that HDP struggles to generate effective trajectories towards the prompt when it does not align with the distribution of demonstrated trajectories.
	We speculate that the reason for the limited guidance is that the lack of demonstration data makes HDP unable to generate effective data for unlearned conditions.
	We will try to improve the controllability of HDP by training it on larger datasets, and explore ways to improve its generalization with limited datasets.
	Secondly, since objective contact guides the output of the Actor, the prediction accuracy of Guider significantly influences the quality of HDP's final outputs. 
	Considering the impressive performance of Visual Language Models (VLMs) in recent years in fields such as logical reasoning and spatial modeling, modeling objective contacts using VLMs is expected to enhance Guider's generalization and prediction accuracy.
	Last, the general formulation of HDP allows it to scale across multiple robots and end-effectors with varying numbers of contact points. 
	We will validate HDP's performance in operational tasks involving multi-robot and multi-finger end-effectors in future work.

	\section{Conclusion}
	In this paper, we propose a manipulation trajectory generation method called Hierarchical Diffusion Policy (HDP) which combines behavior cloning and Q-learning strategy for hierarchical learning.
	Through comprehensive evaluation on four simulated tasks and two real-world tasks, we show that HDP significantly outperforms state-of-the-art imitation learning methods and exhibits greater interpretability and controllability, previously unseen in similar methods, which enables HDP to receive human prompts to generate customized trajectories.
	Our results also validate some key design choices, including phased objective contacts, 3D vision encoders, and snapshot gradient optimization, that are critical to unlocking the full potential of HDP.
	Furthermore, results in real-world scenarios validate that HDP can rely on human prompts to overcome local optima traps and reliably complete tasks under various visual and physical interferences.
	However, the controllability and generalization of HDP are still limited by the scale of training data and model capacity, which will be further explored in our future work.

	\section*{Appendix}
	
	\subsection{Hyperparameters}
	Hyerparameters used for Hierarchical Diffusion Policy on tasks and networks are shown in Tab.~\ref{tab_hyperparameters_task} and Tab.~\ref{tab_hyperparameters_net}.
	The control mode adopts posture control based on the experimental results of comparing posture and velocity control by Difffusion Policy \cite{chi2023diffusion}.
	For the Push-T task where the end-effector is a centrally symmetric circle, the controlled degrees of freedom only include translation in the two-dimensional plane.
	
	HDP is not sensitive to $T_r$ (Fig.~\ref{fig_T_r}), so we keep $T_r$ consistent with $T_a$ for convenience.
	$\tau_l$ and $\tau_r$ are set according to the MRLNM \cite{wang2024multi} to align with the average magnitude of the action.
	In Push-T and Move-T tasks that require precise operations, $\tau_l$ and $\tau_r$ are an order of magnitude smaller than the average amplitude of the action.
	$\tau_f$ is set to one-fourth of the end-effector size to ensure the generated trajectory is accurately aligned with the objective contact while avoiding instability from overly strict thresholds during training.
	
	The objective contact representation includes the contact position of each finger and a flag indicating whether each finger is in contact. 
	The objective contact positions of fingers that are not in contact are all set to 0.
	For the cloth unfolding task involving deformable object, since Algorithm \ref{RecordFunction} struggles to model the object and accurately compute objective contacts, we record the contact position between the end-effector and the object while collecting demonstrations and takes the immediate contact as the objective.
	The actions output by the Actor include the pose and closure state (if any) of the end-effector.
	
	To ensure training stability, we use exponential moving average (EMA) to smooth the weights of the Actor and Guider, as follows:
	\begin{align}
		\theta' = (1-(1+s)^{-\sigma})\theta' + (1+s)^{-\sigma}\theta
	\end{align}
	where $\theta'$ is the smoothed network weight, $s$ is the number of training iterations, and $\sigma$ is used to calculate the smoothing factor.
	
	The weights of the target critic $\phi'$ are obtained by calculating the EMA of the critic weights $\phi$, as follows:
	\begin{align}
		\phi' = (1-\sigma)\phi' + \sigma\phi
	\end{align}
	We maintain two target critic networks concurrently and select the minimum output to compute the target Q-value (Eq.~\ref{eq_49}), enhancing training stability and convergence speed.

	\subsection{Prompt Specifying Method}
\begin{figure}[tp]
		\centering
		{\includegraphics[scale=0.5]{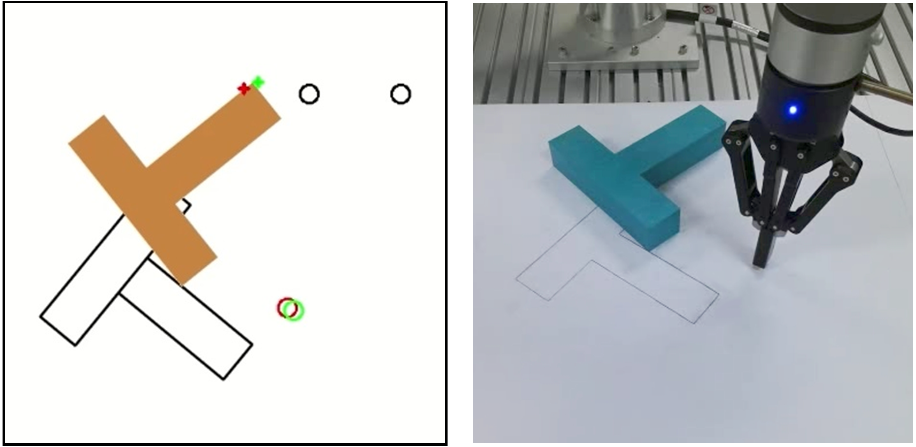}}
		\caption{
			\textbf{Visual interface of Move-T task}.
			The red and green circles indicate the positions of the two fingers of the gripper, and the crosses indicate the prompts.
		}
		\label{fig_prompt_visual}
	\end{figure}

	In this paper, the prompt guidance experiments conducted on the Push-T and Move-T tasks require obtaining the location of the specified prompt through a visual interface.
	The visual interface in the Push-T task is shown in Fig.~\ref{fig_control}. 
	We set the prompt by detecting the mouse click position.
	The visual interface in the Move-T task is shown in Fig.~\ref{fig_prompt_visual}.
	We set the x and y coordinates of the prompt to the mouse click position., and the z coordinate to a constant, which is the mean z coordinate of the demonstrations we collected.

	\section*{Acknowledgment}
	This work is supported by National Natural Science Foundation of China (NO. U22A2058, 62176138, 62176136), National Key R\&D Program of China (NO.2018YFB1305300), Shandong Provincial Key Research and Development Program (NO. 2019JZZY010130, 2020CXGC010207).
	
	\bibliographystyle{Bibliography/IEEEtranTIE}
	\bibliography{Bibliography/IEEEabrv, Bibliography/reference}\

\end{document}